\documentclass[journal]{IEEEtran}
\usepackage{cite}

\usepackage{color,soul}
\definecolor{FrontalColor}{RGB}{251,224,205}
\definecolor{TemporalColor}{RGB}{240,195,195}
\definecolor{ParietalColor}{RGB}{229,241,221}
\definecolor{OccipitalColor}{RGB}{231,230,230}
\usepackage{graphicx}
\usepackage{multirow}
\usepackage[table,xcdraw]{xcolor}

\usepackage{xcolor}
\definecolor{sdgray}{gray}{0.45}
\newcommand{\msd}[2]{\ensuremath{#1\,{\color{sdgray}\scriptstyle(\pm #2)}}}

\usepackage{CJKutf8}

\usepackage{booktabs}

\usepackage{amsmath}
\usepackage{amsthm}
\usepackage{amssymb}

\usepackage{siunitx}
\usepackage{array}
\usepackage{url}
\usepackage{placeins}
\usepackage[caption=false,font=normalsize,labelfont=sf,textfont=sf]{subfig}
\usepackage{algorithm}
\usepackage{algorithmicx} % Provides the framework
\usepackage{algpseudocode} % The specific style package
\usepackage{amsmath}
\usepackage{amssymb}

\ifCLASSINFOpdf
\else
\fi
\usepackage[normalem]{ulem} % normalem 选项防止 \emph 被修改, 提供 \sout (删除线)
\usepackage{xcolor}      % 提供颜色支持

\begin{document}
	\begin{CJK}{UTF8}{gbsn}
		\title{PATH: Preference-Aware Traversability Estimation for Robotic Teleoperation in Off-Road Navigation}
		
		
		\author{Xianghao~Meng$^1$,
			Wenshuo~Wang$^1$,~\IEEEmembership{Member,~IEEE,}
			Cheng~Shao$^1$,
            Lewei~Feng$^1$,
            Qingxiao~Liu$^2$,
            Canyu~Wang$^3$,
			and~Junqiang~Xi$^1$
			\thanks{This work was supported by the National Natural Science Foundation of China (Grant No.: 52595714 and 52572469).  Corresponding Authors: \textit{Wenshuo Wang and Junqiang Xi}}
			\thanks{$^1$X. Meng, W. Wang, C. Shao, L. Feng, and J. Xi are with the School of Mechanical Engineering, Beijing Institute of Technology, Beijing, China, 100081.
			{\tt\small mengxianghao1010@bit.edu.cn, ws.wang@bit.edu.cn, shaocheng@bit.edu.cn, f99600339@163.com, xijunqiang@bit.edu.cn}.}
            \thanks{$^2$Q. Liu is with the Hong Kong Polytechnic University, 11 Yuk Choi Road, Hung Hom, Kowloon, Hong Kong, 999077.
            {\tt\small qingxiao.liu@polyu.edu.hk}.}
            \thanks{$^3$C. Wang is with is with the China North Vehicle Research Institute, Bejing, China, 100072.
            {\tt\small wangcanyu\_bit@163.com}.}
		}

		%\markboth{Journal of \LaTeX\ Class Files,~Vol.~14, No.~8, August~2015}{Shell \MakeLowercase{\textit{et al.}}: Bare Demo of IEEEtran.cls for IEEE Journals}

		\maketitle
		
\begin{abstract}
Traversability estimation is critical for ground vehicle teleoperation, yet existing methods often neglect the human operator's preferences for speed and risk tolerance. To address this gap, we propose PATH-Net, a Preference-Aware Traversability estimation Hierarchical Network that adaptively generates preference-conditioned traversability cost maps and speed profiles for off-road teleoperation. Our framework integrates visual and geometric bird's-eye-view features with short-term historical vehicle states through a multi-head shared-backbone architecture. Our PATH-Net infers traversability cost maps while optimizing speed recommendations by using maximum entropy inverse reinforcement learning and sparse speed supervision. Our method incorporates operator preference via an asymmetric feature aggregation mechanism, splitting inference into a \textit{cost} branch that prioritizes high-risk traversability regions and a \textit{speed} branch that outputs preference-adjusted recommended speeds. Experiments with 15 operators across 225 teleoperated trajectories show alignment with user-specific navigation strategies, achieving modified Hausdorff distance of $0.17\,\mathrm{m}$ (aggressive) and $0.15\,\mathrm{m}$ (conservative). In the questionnaire assessment, participants reported lower NASA-TLX workload scores with PATH-Net recommendations than without recommendations, decreasing from  $34.7$ to $22.2$. These results support preference-conditioned trajectory and speed recommendation for off-road teleoperation.

\end{abstract}

\begin{IEEEkeywords}
    Traversability, Inverse Reinforcement Learning, Off-Road Vehicles, Risk-Aware Planning.
\end{IEEEkeywords}
		
		% \fi  
		\IEEEpeerreviewmaketitle

		%%%%%%%%%%%%%%%%%%%%%%%%%%%%%%%%%%%%%%%%%%%%%%%%%%%%%%%%%%%%%%%%%%%%%%%%
		%%%%%%%%%%%%%%%%%%%%%%%%%%%%%%%%%%%%%%%%%%%%%%%%%%%%%%%%%%%%%%%%%%%%%%%%
\section{Introduction}
\IEEEPARstart{U}{nmanned} ground vehicles (UGVs) have become indispensable tools for exploration and inspection in hazardous environments, where direct human presence is impractical or unsafe (e.g., nuclear facilities, disaster zones) \cite{dinelli2023configurations}. While teleoperation remains the dominant control paradigm for navigating complex unstructured terrain, its effectiveness is fundamentally constrained in non-line-of-sight (NLOS) scenarios. Human operators in such settings must synthesize partial information streams, limited to delayed visual feeds and coarse vehicle-state telemetry \cite{li2023classification}, while deprived of critical physical cues like vestibular feedback and tactile vibrations that enable instinctive control in manned vehicles. This sensory mismatch induces high cognitive load, frequently manifesting as compromised depth perception and degraded situational awareness \cite{du2024sensory}, which collectively elevate the risk of erroneous maneuvering decisions. To mitigate these limitations, there is growing recognition that teleoperation frameworks must evolve beyond pure remote control toward context-aware decision-support systems capable of augmenting human perception and judgment.

Off-road traversability assessment has evolved through multimodal sensor fusion \cite{benrabah2024review}. Pioneering methods primarily utilized exteroceptive geometric information (e.g., digital elevation and 3D LiDAR point clouds) to extract topographic features such as local gradients, discontinuity heights, and roughness metric \cite{asvadi20163d, liang2025gnd}. While effective for structural hazards, these geometric approaches face inherent limitations when discriminating terrains with analogous structures \cite{paton2020navigation, leininger2024gaussian} but different mechanical properties such as soft grass and solid rock. To address this, vision-based semantic segmentation emerged as a complementary solution \cite{jeon2024follow, frey2024roadrunner}. Nevertheless, most supervised learning paradigms demand labor-intensive pixel-wise labeling, which constrains their scalability in unstructured environments \cite{zhang2024autonomous, wang2024traversability}. Recent advances adopt self-supervision by exploiting proprioceptive signals \cite{e2025terrain, bhosale2026review}, including inertial measurements, wheel-terrain interaction forces, and suspension dynamics, as implicit traversability indicators \cite{castro2023does, bae2023self, jung2024v, sivaprakasam2025salon}. Concurrently, visual foundation models like DINOv2 offer pre-trained feature embeddings that mitigate annotation dependency, demonstrating remarkable transferability across off-road domains \cite{oquab2023dinov2, triest2024velociraptor, englert2025vfm}.

\begin{figure}[t]
	\centering
	\includegraphics[width=0.8\linewidth]{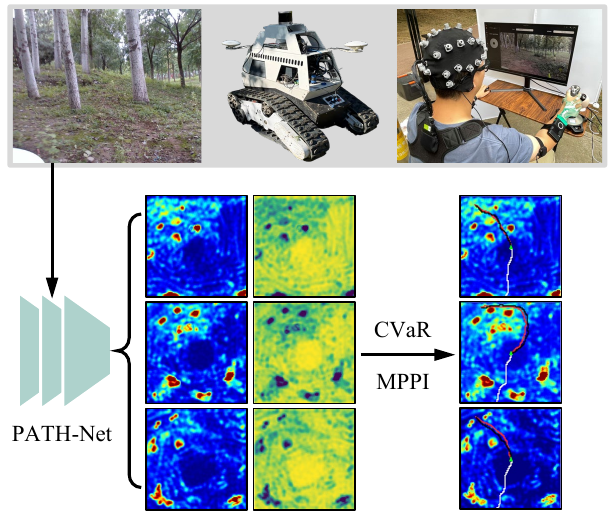}
	\caption{Our PATH-Net integrates environmental factors, kinematics, and expert preferences to provide off-road traversability recommendations.}
    \label{fig:figure01_frame}
\end{figure}

The robotic traversability prediction undergoes a paradigm shift alongside the diversification of sensory data modalities \cite{borges2022survey, dixit2024step, bhosale2026review}. Early approaches relies on deterministic heuristics, where hand-crafted functions mapped geometric features to traversability scores \cite{fan2021learning}. Recent methods have embraced data-driven learning frameworks, evolving along three research thrusts: (i) continuous cost representations learning, (ii)navigation behavior imitation, and (iii) policy optimization from sensory observations \cite{fusaro2026learning}. Representation learning methods such as HDIF (How Does It Feel?) \cite{castro2023does} and Velociraptor \cite{triest2024velociraptor} extract traversability features from visual-geometric inputs, with some architectures additionally predicting speed constraints. Notably, these approaches address a critical limitation of classical planners by capturing terrain properties that defy pure geometric characterization. Complementarily, inverse reinforcement learning (IRL) provides a principled framework for cost function inference. Maximum entropy IRL and its extensions treat demonstrations as outcomes of near-optimal policies, enabling the recovery of implicit cost functions that explain observed navigation behaviors. This proves particularly valuable for encoding human operator preferences and safety considerations that remain opaque to geometry-based planners \cite{zhang2018integrating, triest2023learning}.

Despite these advancements, directly applying traversability assessment methods tailored for autonomous systems to teleoperated off-road vehicles has two limitations. First, they are predominantly operator-agnostic and platform-centric \cite{fankhauser2018probabilistic, sevastopoulos2022survey}, optimizing for single, fixed traversability metrics (e.g., energy efficiency or vibration minimization) while neglecting operator preference. Although terrain--vehicle interaction determines the physical feasibility of traversal, operators may still choose different routes and speed profiles for the same environment \cite{lin2021long, cladera2025air}. For instance, a route deemed optimal by algorithms may  prove unacceptable to risk-averse operators due to their perceived safety differences. This misalignment between system recommendations and human judgment can erode trust \cite{shen2024towards} and trigger automation rejection \cite{hoc2009cooperation}. Second, many approaches rely on static environmental representations, failing to incorporate dynamic vehicle-state context \cite{zhang2018integrating}. Vehicle motion states such as speed transients, angular kinematics, roll/pitch dynamics could influence terrain negotiation strategies for human operators. Therefore, historical motion contains complementary information beyond single-frame terrain analysis, particularly for operator's adjustment anticipation.

In this paper, we propose PATH-Net (a Preference-Aware Traversability estimation Hierarchical Network) for robotic teleoperation, which directly learns candidates of cost functions and speed maps from operator demonstrations. Unlike traversability models that produce a single cost function, our PATH-Net leverages a questionnaire-derived preference parameter to asymmetrically aggregate these candidates, emphasizing upper-tail costs or selecting recommended speed profiles according to the operator preference. This design allows one trained model to generate aggressive, neutral, and conservative navigation advice without training separate models for various operators. This paper's contributions are as follows:
\begin{enumerate}
    \item We introduce a spatiotemporal-BEV temporal convolutional network to dynamically couple terrain perception with vehicle motion history, enabling interactive traversability prediction and yielding kinematic-specific costs and speeds.

    \item Our multi-head network with entropy-regularized IRL unifies aggressive/conservative demonstrations through shared feature extraction and parallel prediction of cost functions and speed profiles. %This eliminates separate model requirement for different behaviors, simplifying deployment while maintaining diversity.
    
	\item We introduce a risk-aware approach that employs conditional value-at-risk (CVaR) optimization for upper-tail cost aggregation and quantile-based speed selection, enabling explicit trade-offs between risk and efficiency. A single parameter controls both the conservatism of the cost map via CVaR's risk threshold and the recommended speed bounds via quantile selection, generating interpretable behaviors ranging from aggressive to cautious. %This unified approach offers tunable risk sensitivity while maintaining computational efficiency.
\end{enumerate}

The remainder of this paper is organized as follows. Section~II formulates the teleoperation decision-support problem and provides an overview of PATH-Net. Section~III presents the feature extraction, cost and speed learning, and asymmetric aggregation methods. Section~IV describes the experimental platform, dataset, and evaluation protocol. Section~V reports the experimental results, and Section~VI concludes the paper.

\section{Problem Formulation}
\label{sec:framework}
%Rather than directly replacing the human operator, PATH-Net generates preference-conditioned cost and speed maps for a planner to produce advisory trajectories. %We first define the problem and then presents a high-level overview of our framework's architecture and core principles.

\begin{figure*}[t]
	\centering
	\includegraphics[width=1.0\linewidth]{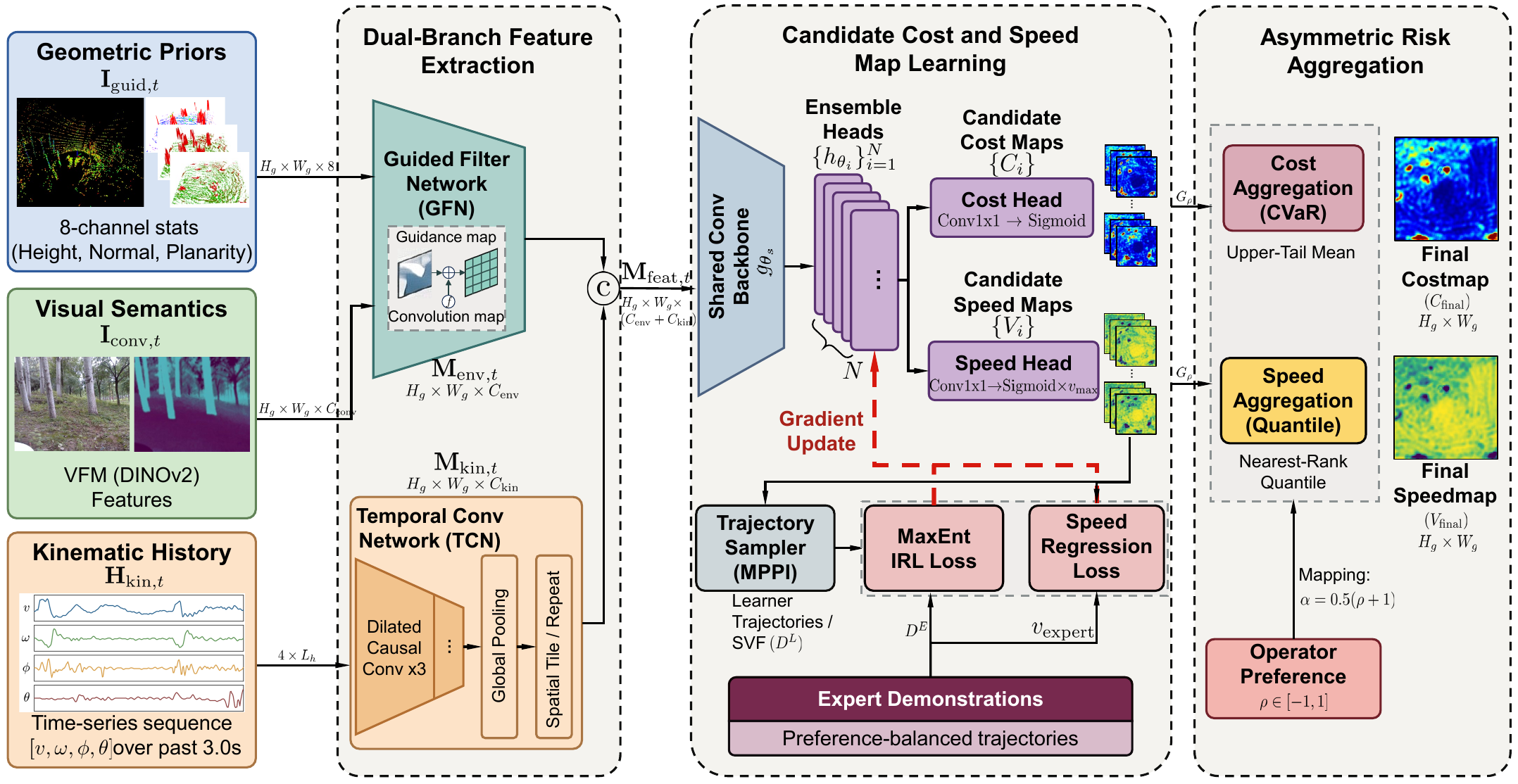}
	\caption{Overview of PATH-Net. Visual-geometric BEV features and recent vehicle states history are fused to predict candidate cost and speed maps, which are then converted into preference-conditioned cost and speed maps by upper-tail cost aggregation and quantile-based speed selection.}
    \label{fig:path_net_architecture}
\end{figure*}

\subsection{Problem Definition}
\label{sec:problem_definition}
We address the problem of non-line-of-sight teleoperation for off-road navigation, where an operator controls a UGV in unstructured terrain with limited sensory feedback. In such scenarios, traversability depends not only on terrain geometry but also on operator preferences. For instance, an efficiency-preferred operator may favor shorter, faster paths, while a risk-averse operator may prioritize slower trajectories with greater obstacle clearance. To accommodate these differences, PATH-Net aims to generate recommendations that reflect both terrain traversability and the operator's route-speed preference.

To generate preference-conditioned recommendations, the state representation should include both environmental perception and recent vehicle-state context. We represent the local environment as a BEV feature map $\mathbf{M}_{\mathrm{env},t}$ that jointly encodes visual appearance and geometric terrain properties. As off-road traversability depends on both terrain characteristics and vehicle dynamics, we augment this representation with a kinematic history vector,
\begin{equation} 
\mathbf{H}_{\mathrm{kin},t} = [\boldsymbol{\xi}_{t-L_h+1},\ldots,\boldsymbol{\xi}_{t}], \ \mathrm{with} \ \boldsymbol{\xi}_{k} = [v_k,\omega_k,\phi_k,\theta_k]^{\top},
\end{equation}
where $v_k$ is the linear velocity, $\omega_k$ the angular velocity, $\phi_k$ the roll angle, and $\theta_k$ the pitch angle at timestep $k$. The state used for prediction is
\begin{equation}
S_t = (\mathbf{M}_{\mathrm{env},t},\mathbf{H}_{\mathrm{kin},t}).
\end{equation}
Here, the parentheses denote a composite state composed of two heterogeneous components, rather than vector concatenation.

The teleoperated demonstrations are collected from expert participants in identical off-road conditions. These demonstrations represent diverse trade-offs between route efficiency and vehicle safety, rather than a single optimal traversability solution. The dataset is defined as

\begin{equation}
\mathcal{D}_{E} = \{(\tau_m,\rho_{u(m)})\}_{m=1}^{M},
\end{equation}
where $\tau_m$ is the $m$-th demonstrated trajectory, $u(m)$ indicates the corresponding participant, and $\rho_{u(m)}\in[-1,1]$ is the participant-specific preference parameter obtained from pre-experiment questionnaires.

Given the current state and a desired preference parameter, the teleoperation support system should generate both spatial guidance (i.e., where to move) and speed recommendations (i.e., how fast to move). Our PATH-Net outputs two BEV-aligned maps: a traversal costmap $C_{\mathrm{final}}\in[0,1]$ indicating preferred routes, and a speed map $V_{\mathrm{final}}\in[0,v_{\max}]$ specifying local speed limits. These maps provide a planner-compatible representation of \textit{preference-aware} traversability.

Instead of direct preference injection, our PATH-Net decouples terrain analysis from preference adaptation. First, we utilize a prediction network to generate a set of candidate cost and speed maps,
\begin{equation}
\{(C_i,V_i)\}_{i=1}^{N} = F_{\theta}(S_t).
\end{equation}

These candidates represent alternative interpretations of the terrain and vehicle state. The operator preference is incorporated during aggregation:
\begin{equation}
(C_{\mathrm{final}},V_{\mathrm{final}}) = G_{\rho}\left(\{(C_i,V_i)\}_{i=1}^{N}\right).
\end{equation}
where $G_{\rho}$ denotes the aggregation operator in Fig. \ref{fig:path_net_architecture}, consisting of upper-tail cost aggregation and quantile-based speed selection. $\rho$ is a questionnaire-derived preference parameter and is not provided to the prediction network. It affects the outputs only through $G_{\rho}$: smaller values of $\rho$ correspond to efficiency-oriented operation and select higher recommended speeds, whereas larger values correspond to more cautious operation by emphasizing higher-cost candidates and selecting lower recommended speeds.

\begin{figure*}[t]
	\centering
	\includegraphics[width=0.8\linewidth]{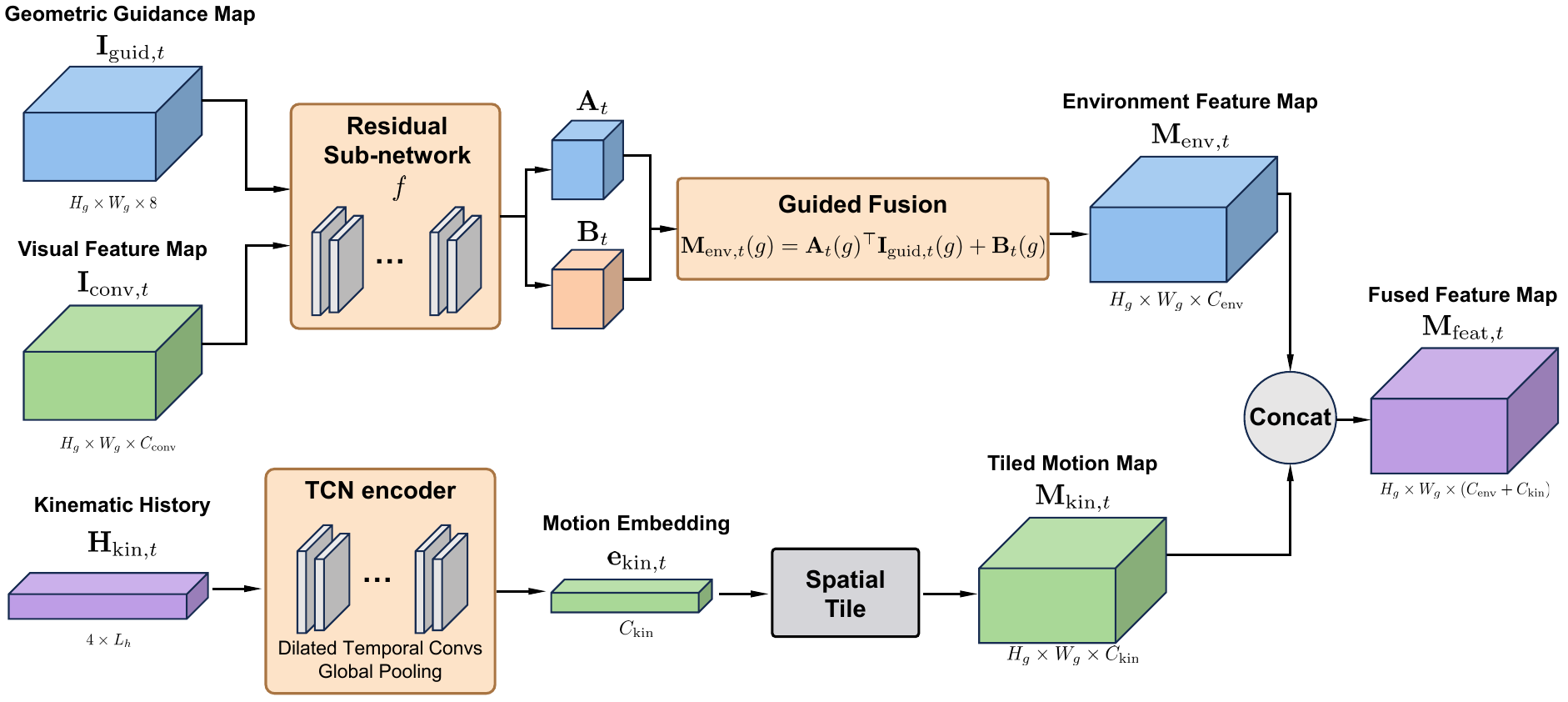}
	\caption{The spatiotemporal feature fusion module. The geometric guidance map is fused with visual features to form the environment feature map, while the kinematic history is encoded by a TCN and spatially tiled. Their concatenation yields the fused feature map used for subsequent cost and speed prediction.}
    \label{fig:feature_fusion_detail}
\end{figure*}

\subsection{Framework Overview}
\label{sec:framework_overview}
Our proposed PATH-Net architecture (Fig.~\ref{fig:path_net_architecture}) processes two complementary sensory inputs. The environmental branch generates a BEV representation, $\mathbf{M}_{\mathrm{env},t}$, by fusing LiDAR geometry and camera features through a Guided Filter Network (GFN). Simultaneously, a Temporal Convolutional Network (TCN) encodes the kinematic history, $\mathbf{H}_{\mathrm{kin},t}$. These representations are concatenated to form the joint feature map, $\mathbf{M}_{\mathrm{feat},t}$. This joint feature map feeds into a convolutional backbone with parallel prediction heads. Each head outputs candidate cost maps and speed maps. The cost map is learned via Maximum Entropy IRL to match human operator preferences, while the speed map is learned from speed supervision. These heads generate diverse candidate cost and speed maps for subsequent aggregation without explicit preference specialization.

During inference, the questionnaire-derived preference parameter $\rho$ is mapped to an aggregation level $\alpha$. PATH-Net then aggregates the candidate maps asymmetrically: the cost maps use upper-tail aggregation, while the speed maps use quantile-based selection to generate preference-dependent speed bounds. The final outputs, $C_{\mathrm{final}}$ and $V_{\mathrm{final}}$, guide the path planner to generate the recommended trajectory.

\section{The Proposed PATH-Net}
\label{Method}
PATH-Net comprises three components (Fig.~\ref{fig:path_net_architecture}): a dual-branch feature extraction module capturing environmental and vehicle dynamic motion features, a multi-head shared-backbone network generating candidate cost and speed maps, and an asymmetric aggregation module producing preference-conditioned recommended maps.

\subsection{Dual-Branch Feature Extraction}
\label{sec:feature_extraction}
The feature extraction module contains an environmental branch and a temporal branch. Building on the guided fusion approach in~\cite{jeon2024follow}, we use GFN  to combine geometric and visual information for the environmental branch, and use TCN to encode recent vehicle-motion and attitude measurements for the temporal branch. A detailed view of the spatiotemporal feature fusion module is shown in Fig.~\ref{fig:feature_fusion_detail}. The figure illustrates how visual features guide the transformation of geometric BEV features and how the kinematic history is encoded and aligned with the BEV grid before channel-wise concatenation.

\subsubsection{Environmental Perception and Fusion}
This module generates a unified environmental feature map, $\mathbf{M}_{\mathrm{env},t} \in \mathbb{R}^{H_g\times W_g\times C_{\mathrm{env}}}$, through multi-sensors fusion. We employ GFN \cite{jia2021residual, zhang2022guided} with 4 residual blocks and a guided filter layer to integrate geometric and visual features.

\textbf{Guidance Image}, $\mathbf{I}_{\text{guid}}$, is an eight-channel BEV representation constructed from LiDAR point clouds. It includes three height statistics, planarity and linearity from the local point covariance, and three surface-normal components for each grid cell. These channels provide local shape and terrain-structure cues that guide the subsequent fusion with visual features.

\textbf{Visual Feature Map}, $\mathbf{I}_{\mathrm{conv}}$, is extracted from camera images using a pre-trained DINOv2 model \cite{oquab2023dinov2}, encoding terrain appearance  onto the BEV grid. This provides appearance cues for terrain regions that may have similar geometric shapes but different surface properties, while reducing the need for dense pixel-level annotation.

\textbf{GFN-Guided Fusion}. The relation between visual appearance and geometric traversability is not uniform over the BEV grid. For example, a visually vegetated region and a compact soil region may have similar local height profiles but different traversal properties. Therefore, we use the visual feature map to generate cell-wise fusion coefficients:

\begin{equation}
\left( \mathbf{A}_t,\mathbf{B}_t \right) = f_{\mathrm{res}} \left( \mathbf{I}_{\mathrm{conv},t} \right). \label{eq:gfn_coefficients}
\end{equation}
For each grid cell $g$, we have $\mathbf{A}_t(g) \in \mathbb{R}^{C_{\mathrm{guid}}\times C_{\mathrm{env}}}, \mathbf{B}_t(g) \in \mathbb{R}^{C_{\mathrm{env}}}$, where $C_{\mathrm{guid}}$ is the number of geometric guidance channels. The fused environmental feature is computed as

\begin{equation}
\mathbf{M}_{\mathrm{env},t}(g) = \mathbf{A}_t(g)^{\top} \mathbf{I}_{\mathrm{guid},t}(g) + \mathbf{B}_t(g).
\label{eq:gfn_fusion}
\end{equation}
In this fusion, the visual feature at each grid cell controls how the corresponding LiDAR geometric vector is transformed. The fused feature $\mathbf{M}_{\mathrm{env},t}$ therefore keeps the BEV structure of the geometric priors while adding appearance-dependent terrain information.

\subsubsection{Kinematic State Encoding}
We employ TCN to encode the recent kinematic history of the vehicle. The input sequence comprises $L_h$ historical frames

\begin{equation} 
\mathbf{H}_{\mathrm{kin},t} = \left[ \boldsymbol{\xi}_{t-L_h+1}, \ldots, \boldsymbol{\xi}_{t} \right], 
\label{eq:kinematic_input} 
\end{equation} 
where $L_h=30$ (equivalent to $3.0\,\mathrm{s}$ at $10\,\mathrm{Hz}$). The TCN architecture utilizes dilated causal convolutions to ensure temporal causality, with each layer's output depending only on present and past inputs. At each layer $\ell$, convolutional weights are normalized using 

\begin{equation} 
\overline{\mathbf{W}}^{(\ell)} = \operatorname{WeightNorm} \left( \mathbf{W}^{(\ell)} \right), 
\label{eq:tcn_weight_norm} 
\end{equation} 
The layer-wise computation is given by

\begin{equation} 
\mathbf{h}_{t}^{(\ell)} = \operatorname{ReLU} \left( \sum_{r=0}^{K_c-1} \overline{\mathbf{W}}_{r}^{(\ell)} \mathbf{h}_{t-d_{\ell}r}^{(\ell-1)} + \mathbf{b}^{(\ell)} \right), 
\label{eq:tcn_layer} 
\end{equation} 
where $K_c$ is the temporal kernel size and $d_{\ell}$ is the dilation factor at layer $\ell$. The final kinematic embedding $\mathbf{e}_{\mathrm{kin},t} \in \mathbb{R}^{C_{\mathrm{kin}}}$ is obtained via global average pooling.

Unlike a dynamic model, the TCN encoder is used only as a history encoder rather than a future-state predictor. It summarizes recent speed changes, angular-rate variations, and roll/pitch responses, so that the subsequent cost and speed predictions can account for motion-dependent traversability.
		
\subsubsection{Feature Fusion}
The kinematic embedding $\mathbf{e}_{\mathrm{kin},t}$ summarizes the recent vehicle-state context but has no spatial dimension. To condition each BEV cell on the same motion context, we spatially tile this embedding to match the environmental feature map:
\begin{equation} 
\mathbf{M}_{\mathrm{kin},t} = \operatorname{Tile} \left( \mathbf{e}_{\mathrm{kin},t}, H_g, W_g \right). 
\label{eq:kinematic_tiling} 
\end{equation} 
The final fused feature representation is then constructed through channel-wise concatenation

\begin{equation} 
\mathbf{M}_{\mathrm{feat},t} = \operatorname{Concat} \left( \mathbf{M}_{\mathrm{env},t}, \mathbf{M}_{\mathrm{kin},t} \right). 
\label{eq:feature_fusion} 
\end{equation}
This concatenation lets the shared backbone estimate cost and speed from local terrain features together with the recent vehicle-motion context at each BEV cell.

\subsection{Multi-Head Candidate Cost and Speed Map Learning}
\label{sec:policy_learning}
Given the feature map $\mathbf{M}_{\mathrm{feat},t}$, our model predicts a set of pairs of normalized cost and speed maps for planning. Using a shared  convolutional backbone $g_{\theta_s}$ and specialized heads $\{h_{\theta_i}\}_{i=1}^{N}$, each head generates 

\begin{equation}
\left( C_i,V_i \right) = h_{\theta_i} \left( g_{\theta_s} \left( \mathbf{M}_{\mathrm{feat},t} \right) \right), i=1,\ldots,N. 
\label{eq:multi_head_output} 
\end{equation} 
Notably, individual heads are not pre-assigned to specific operator preferences. Instead, they generate diverse candidate maps for subsequent preference-conditioned aggregation \cite{nan2025safe}. To encourage diversity among the output heads, all heads are initialized independently, and each head is trained using a stratified bootstrap sample drawn from a preference-balanced mini-batch. The training protocol follows two update rules: The shared backbone is updated using the average loss across all heads, and each head is updated through its assigned bootstrap sample. Notably, preference labels are used only for data sampling and remain absent from network inputs.

\subsubsection{Output Parameterization and Normalization}
Each output head generates two logit maps: a cost logit map $Z_i^{C}$ and a speed logit map $Z_i^{V}$. The cost logits are mapped to a bounded range by a sigmoid activation, so that the resulting costmap can be used as a normalized relative traversal cost for aggregation and planning:
\begin{equation} 
C_i = \sigma \left( Z_i^{C} \right) \in [0,1]^{H_g\times W_g}, 
\label{eq:cost_normalization} 
\end{equation} 
where $C_{i}$ values represent relative traversal costs (lower values indicate preferable regions). This dimensionless quantity serves as a unified metric for planning and evaluation.

The speed prediction is similarly normalized but scaled to physical units:

\begin{equation} 
V_i = v_{\max} \sigma \left( Z_i^{V} \right) \in [0,v_{\max}]^{H_g\times W_g}.
\label{eq:speed_normalization} 
\end{equation}
where $v_{\max}$ is determined by the maximum observed speed in the teleoperation dataset (here, $v_{\max}=1.2\,\mathrm{m/s}$). During training, speed annotations are clipped to $[0, v_{\max}]$ and normalized by this value. All reported speed errors are in original physical units (m/s). Note that the normalization range is preference-invariant; operator preference influence emerges solely through the aggregation operator (Section~\ref{sec:asymmetric_aggregation}). The final speed map provides recommended speed limits for motion planning, not absolute safety boundaries. 

\subsubsection{Cost Map Learning}
The cost map learning process adopts the maximum entropy IRL framework \cite{zhang2018integrating,triest2023learning}. For each head, let $D^{E}(g)$ denote the normalized expert state visitation frequency over the BEV grid, and $D_i^{L}(g)$ the corresponding learner visitation frequency induced by the cost map $C_i$. The cost map gradient is given by  

\begin{equation} 
\frac{\partial \mathcal{L}_{\mathrm{IRL}}^{(i)}} {\partial C_i(g)} = D^{E}(g) - D_i^{L}(g). 
\label{eq:irl_cost_gradient} 
\end{equation} 
The gradient update reduces costs for demonstrator-visited states while increasing costs for regions frequently visited by the learner. With parameters $\Theta_i=\{\theta_s,\theta_i\}$ and $L_{2}$  regularization coefficient $\lambda_{c}$, the complete parameter gradient becomes 

\begin{equation} 
\nabla_{\Theta_i} \left( \mathcal{L} \right) = \sum_{g\in\mathcal{G}} \frac{\partial \mathcal{L}_{\mathrm{IRL}}^{(i)}} {\partial C_i(g)} \nabla_{\Theta_i}C_i(g) + \lambda_c \nabla_{\Theta_i} \lVert C_i\rVert_2^2,
\label{eq:irl_parameter_gradient} 
\end{equation} 
with $\mathcal{L}=\mathcal{L}_{\mathrm{IRL}}^{(i)} + \lambda_c \lVert C_i\rVert_2^2$. We estimate $D_i^{L}(g)$ using MPPI-based trajectory sampling. From $N_r$ rollouts $\{\tau_n\}_{n=1}^{N_r}$ with horizon $H_p$ and importance weight $\eta_n$, the visitation frequency is computed as

\begin{equation} 
D_i^{L}(g) = \frac{1}{\mathcal{Z}_{\mathrm{vis}}} \sum_{n=1}^{N_r} \eta_n \sum_{j=1}^{H_p} \mathbb{I} \left[ \Pi(x_{n,j})=g \right], 
\label{eq:learner_svf} 
\end{equation} 
where $x_{n,j}$ is the $j$-th state in rollout $\tau_n$, $\Pi(\cdot)$ projects continuous vehicle pose to discrete grid cells, and $\mathcal{Z}_{\mathrm{vis}}$ ensures probability normalization. This formulation avoids explicit  discretizations of the vehicle control space.

\subsubsection{Speed Map Learning}
Speed labels are only available at discrete points along the demonstrated trajectory. Given a demonstration 
\[ \tau_E = \left\{ (\mathbf{p_q},v_q) \right\}_{q=1}^{L_{\tau}}, \] 
we define the label-dilation set as 

\begin{equation} 
\Omega_E = \left\{ g\in\mathcal{G} \,\middle|\, \min_{1\leq q\leq L_{\tau}} \lVert \mathbf{c(g)}-\mathbf{p_q}\rVert_2 \leq r_f \right\}, 
\label{eq:speed_label_region} 
\end{equation} 
where $\mathbf{c(g)}$ is the center of grid cell $g$, and $r_f=0.5\,\mathrm{m}$ defines the dilation radius. This operation extends sparse speed labels to nearby grid cells without affecting collision checking. Each labeled cell $g$ is assigned the speed value from its nearest trajectory point

\begin{equation} 
v_E(g) = v_{q^{*}}, \quad q^{*} = \underset{1\leq q\leq L_{\tau}} {\operatorname{arg\,min}} \, \lVert \mathbf{c(g)}-\mathbf{p_q}\rVert_2 . 
\label{eq:nearest_speed_label} 
\end{equation} 
The speed loss for prediction head $i$ is computed as

\begin{equation} 
\mathcal{L}_{\mathrm{speed}}^{(i)} = \frac{1}{|\Omega_E|} \sum_{g\in\Omega_E} \left| \widehat{V}_i(g) - \widehat{v}_E(g) \right|, \label{eq:speed_loss} 
\end{equation} 
with \[ \widehat{V}_i = \frac{V_i}{v_{\max}}, \quad \widehat{v}_E = \frac{v_E}{v_{\max}}. \] 
Despite the sparse supervision in $\Omega_E$, the fully convolutional network produces a dense speed map during inference. For each prediction head $i$ with its stratified bootstrap mini-batch $\mathcal{B}_i$, the complete training objective is 

\begin{equation}
\mathcal{L} = \frac{1}{N} \sum_{i=1}^{N} \left[ \mathcal{L}_{\mathrm{IRL}}^{(i)} \left( \mathcal{B}_i \right) + \lambda_v \mathcal{L}_{\mathrm{speed}}^{(i)} \left( \mathcal{B}_i \right) + \lambda_c \lVert C_i\rVert_2^2 \right]. 
\label{eq:complete_training_objective} 
\end{equation}

\subsection{Preference Adaptation via Asymmetric Aggregation}
\label{sec:asymmetric_aggregation}

The candidate maps are transformed into preference-conditioned recommended maps via a preference parameter $\rho\in[-1,1]$. The parameter is linearly mapped to an aggregation weight $\alpha\in[0,1]$: 

\begin{equation}
\alpha = \frac{\rho+1}{2}. 
\label{eq:style_to_aggregation} 
\end{equation} 
The cost and speed maps employ distinct aggregation operators to preserve their differing semantic meanings. The cost maps use an upper-tail mean, while speed maps adopt a nearest-rank quantile approach. This aggregation operates grid cell-wise across all $N$ candidate maps. Note that $\alpha$ serves an operational aggregation parameter rather than a true probability measure due to finite $N$. In our implementation, we set $N=5$. 

\textbf{Upper-Tail Cost Aggregation.} For each grid cell $g$, the candidate costs are sorted in ascending order 
\begin{equation}
C_{(1)}(g) \leq \cdots \leq C_{(N)}(g). 
\label{eq:sorted_costs}
\end{equation}

To robustly aggregate high-cost outliers, a finite-sample upper-tail mean is computed based on conditional value-at-risk (CVaR) principles~\cite{hong2014monte}. The number of retained candidates is defined as
\begin{equation} 
k_{\alpha} = \max \left( 1, \left\lceil (1-\alpha)N \right\rceil \right),
\label{eq:cost_tail_size}
\end{equation} 
yielding the final cost map
\begin{equation}
C_{\mathrm{final}}(g) = \frac{1}{k_{\alpha}} \sum_{m=N-k_{\alpha}+1}^{N} C_{(m)}(g).
\label{eq:cost_aggregation} 
\end{equation} 

When $N=5$, $\alpha=0.95$ selects the maximum candidate cost, while $\alpha=0.05$ averages all five candidates under the nearest-rank approximation. 

\textbf{Speed Quantile Selection.} The candidate speeds are ascendingly sorted as 

\begin{equation} 
V_{(1)}(g) \leq \cdots \leq V_{(N)}(g). 
\label{eq:sorted_speeds} 
\end{equation} 
The nearest-rank index is computed as
\begin{equation} r_{\alpha} = \operatorname{clip} \left( \left\lceil (1-\alpha)N \right\rceil, 1, N \right), 
\label{eq:speed_quantile_index} 
\end{equation} 
yielding the recommended speed map
\begin{equation}
V_{\mathrm{final}}(g) = V_{(r_{\alpha})}(g). 
\label{eq:speed_aggregation} 
\end{equation}

With $\alpha=(\rho+1)/2$, smaller $\alpha$ values correspond to the aggressive setting and select higher speed candidates, whereas larger $\alpha$ values correspond to the conservative setting and select lower speed candidates. Compared to cost aggregation, this method selects a single quantile instead of averaging the lower tail, thus preventing the accumulation of multiple low-speed predictions. Because cost and speed candidates are sorted independently at each grid cell, the marginal aggregation may break the original pairing between each head's cost and speed predictions.

\section{Experimental Setup and Data Collection}
\label{sec:setup}

\subsection{Participants and Preference Assessment}
\label{sec:participants}
We recruited 15 participants (mean age: 24.0 years, SD = 2.4) for the study. The Institutional Review Board (IRB) of Beijing Institute of Technology approved the experimental protocol, and all participants provided written informed consent after being informed of the procedures, and their right to withdraw. Prior to data collection, each participant completed a structured training session with over 8 hours of teleoperation practice on average.

To evaluate individual differences in risk preferences during remote navigation, we developed a five-item questionnaire based on established Risk-Taking Propensity (RTP) measures in human-machine interaction \cite{matsushita2021operator, rolison2023psychological}. The questionnaire assessed two key dimensions: self-perceived control ability and efficiency-safety tradeoff preferences. Each item was rated on a five-point Likert scale. For items Q1-Q4, higher scores indicate greater self-reported confidence or efficiency preference. Since the item Q5 reflects safety-oriented behavior, its score was inverted as
\begin{equation}
\widetilde{q}_{5}=6-q_{5}.
\label{eq:reverse_score}
\end{equation}

The total score for participant $u$ is given by
\begin{equation} 
R_u=q_1+q_2+q_3+q_4+\widetilde{q}_{5}, R_u\in[5,25]. \label{eq:style_score} 
\end{equation}
where higher values of $R_u$ indicate efficiency-oriented operation, and lower values correspond to conservative operation.

Participants were classified into three groups based on predetermined score thresholds along the five-point scale
\begin{equation}
\kappa_u= \begin{cases} \text{Aggressive}, & R_u\geq18,\\ \text{Neutral}, & 14\leq R_u\leq17,\\ \text{Conservative}, & R_u\leq13. \end{cases}
\label{eq:style_assignment}
\end{equation}

Following this classification, we obtained 4 aggressive, 6 neutral, and 5 conservative participants. Each group was characterized by a preference parameter $\rho$ and corresponding aggregation level $\alpha$, as detailed in Table~\ref{tab:participant_styles}. The mapping from $\rho$ to $\alpha$ is defined in Section~\ref{sec:asymmetric_aggregation}.

\begin{table}[t]
	\centering
	\caption{Questionnaire-Defined Preferences and Aggregation Parameters}
	\label{tab:participant_styles}
	\begin{tabular}{lccc} 
		\hline
		\textbf{Preference Category} & \textbf{Assigned $\rho$} & \shortstack{\textbf{Aggregation}\\\textbf{Level $\alpha$}} & \textbf{Participants} \\ \hline
		Aggressive   & $-0.9$ & $0.05$ & 4 \\
		Neutral      & $0.0$  & $0.50$ & 6 \\
		Conservative & $+0.9$ & $0.95$ & 5 \\ \hline
		\textbf{Total} & - & - & \textbf{15} \\ \hline
	\end{tabular}
\end{table}

\subsection{Experimental Platform and Apparatus}
\label{sec:apparatus}
The experimental platform was a UGV platform equipped with a multimodal sensor suite for off-road teleoperation and data collection. The sensor configuration was designed to provide both exteroceptive environmental perception and vehicle-state measurements required by the proposed feature extraction architecture.

The UGV was equipped with a LiDAR and a Camera. 
The \textbf{LiDAR} is a RoboSense RS-LiDAR-16 sensor, mounted at a height of $0.8$ m, and used to acquire 3-D point cloud data for constructing the geometric priors in $\mathbf{I}_{\mathrm{guid}}$. The \textbf{camera} is an Intel RealSense D435i depth camera and used to capture RGB images with a resolution of $1280 \times 720$. These images were used to compute the visual feature representation $\mathbf{I}_{\mathrm{conv}}$.

The kinematic history $\mathbf{H}_{\mathrm{kin},t}$ was constructed using vehicle-state measurements from a StarNeto Newton-M2 GNSS high-precision Attitude and Heading Reference System (AHRS) integrated with RTK-GPS. The system provided linear velocity $(v_k)$, angular-rate measurement $(\omega_k)$, and roll/pitch angles $(\phi_k,\theta_k)$ at $10$ Hz.

\begin{figure}[t]
	\centering
	\includegraphics[width=1.0\linewidth]{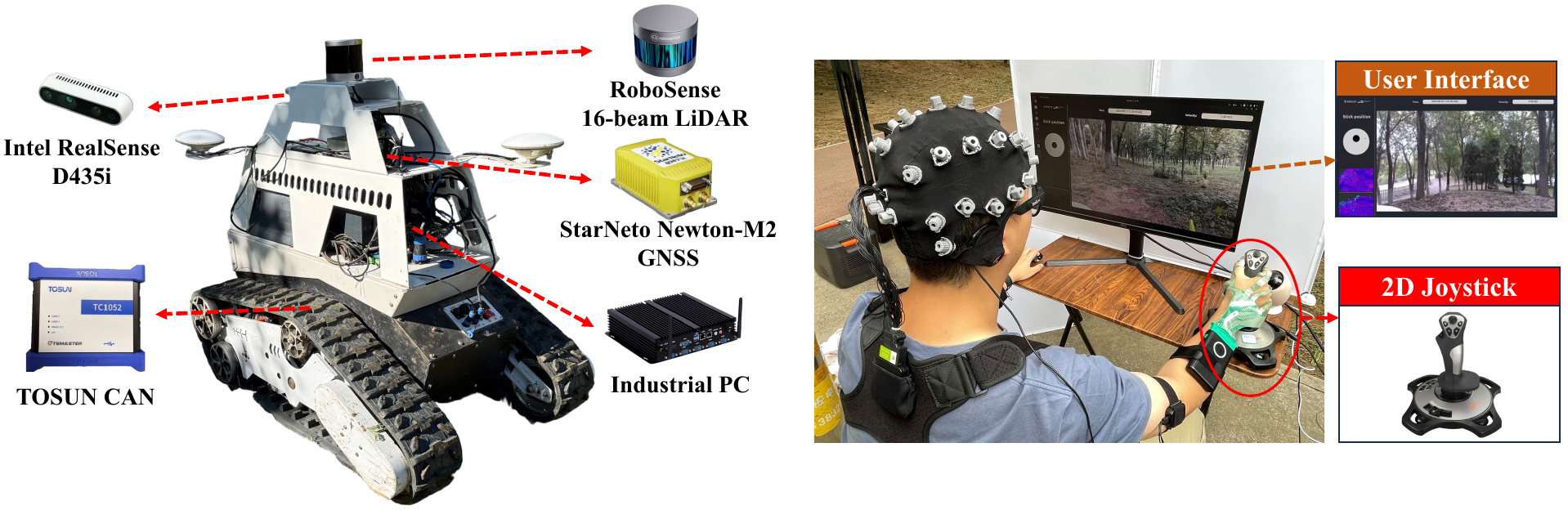} 
	\caption{Experimental hardware and teleoperation system.}
	\label{fig:shebei}
\end{figure}

The teleoperation experiments were conducted under non-line-of-sight conditions from a remote station. The operator interface displayed a front-facing camera feed and real-time vehicle-state data. Control commands were issued via a PXN 2113PRO joystick, with all sensor data and commands synchronized using the Robot Operating System (ROS)~\cite{quigley2009ros} and transmitted through a local 5G wireless network (see Fig.~\ref{fig:shebei}).

\subsection{Scenarios and Data Collection}
\label{sec:scenarios}
Experiments were conducted on unstructured hillside terrain near Beijing. The test site featured varying ground conditions, including slopes with inclinations up to $30^{\circ}$, ditches with depths of $0.1$--$0.2\,\mathrm{m}$, and mixed vegetation density areas.

During each trial, participants remotely operated the UGV between predefined start and goal positions. The terrain configuration allowed for multiple navigation strategies, including longer routes with gentler slopes ($\leq 15^{\circ}$) and more direct routes requiring traversal of steeper terrain ($15^{\circ}\sim 30^{\circ}$). These alternatives allowed participants to express different path and speed preferences during data collection. Fifteen human operators each completed 15 trials, generating a total of 225 unique teleoperated trajectories. Data collection spanned one month to capture varying environmental conditions.

\subsection{Data preprocessing}
\label{sec:preprocessing}
The raw dataset containing 225 trajectories was processed to generate inputs for PATH-Net. At each timestamp $t$, sensor data was projected onto a $10\,\mathrm{m}\times10\,\mathrm{m}$ BEV grid with $0.1\,\mathrm{m}$ resolution (i.e., $100\times100$ cells). 

For $I_{\mathrm{guid}}$, LiDAR point clouds were filtered and projected onto the grid to encode elevation and occupancy information. For $I_{\mathrm{conv}}$, RGB images were processed using a DINOv2-Small backbone to extract visual features. These features were then associated with 3D points using registered depth image from the D435i camera, transformed into the vehicle coordinate frame using the calibrated camera parameters, and projected onto the BEV grid.

Vehicle-state measurements were synchronized and sampled at $10\,\mathrm{Hz}$. The data sequences were segmented into $3.0\,\mathrm{s}$ overlapping windows ($L_h=30$ frames). Demonstrated vehicle speeds were smoothed using a Gaussian kernel with a standard deviation of five frames to remove high-frequency noise from sensor measurements and control inputs, providing cleaner supervision signals for speed map learning.

\section{Result Analysis and Discussion}
\label{sec:experiments}
This section presents three key evaluations of PATH-Net: trajectory alignment, self-reported workload, and the effect of preference settings on generated trajectories. We examine whether the recommended trajectories match held-out demonstrations, whether participants report lower workload with PATH-Net recommendations, and how the preference parameter changes route choice and recommended speed.

\begin{figure}[t]
	\centering
	\includegraphics[width=1.0\linewidth]{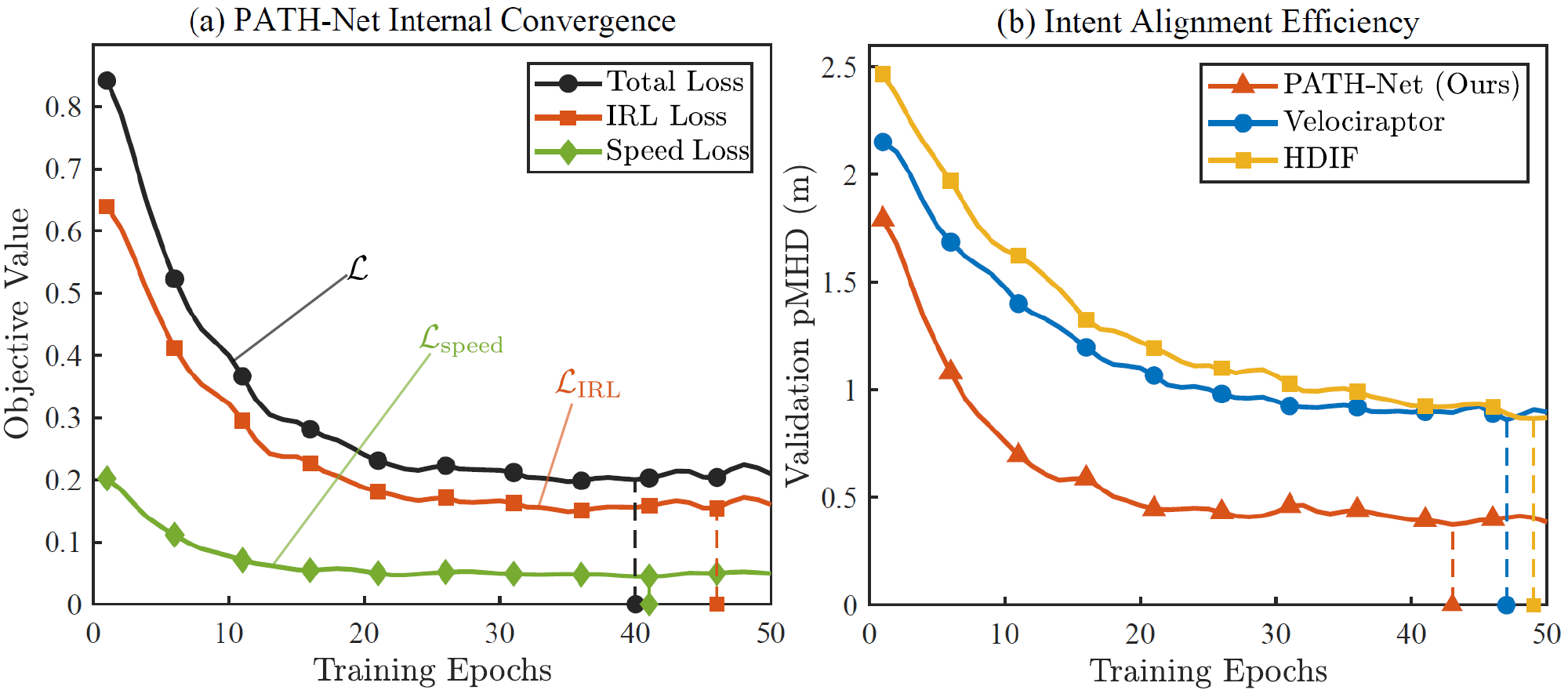} 
    \caption{Training convergence and validation-trajectory alignment. (a) Training objectives of PATH-Net over 50 epochs, including the total objective $\mathcal{L}$, IRL loss $\mathcal{L}_{\mathrm{IRL}}$, and speed regression loss $\mathcal{L}_{\mathrm{speed}}$. (b) Validation pMHD of PATH-Net and the adapted baselines under the same 50-epoch training budget.}
	\label{fig:learning_curves}
\end{figure}

\subsection{Implementation Details}
\label{sec:implementation}
\subsubsection{Dataset Partition}
We adopt a participant-disjoint split strategy to ensure no data leakage between training and evaluation phases. This allocates 9 participants with 135 trajectories for training (2 aggressive, 4 neutral, 3 conservative), 3 participants with 45 trajectories for validation, and 3 participants with 45 trajectories for testing, with the latter two sets each containing one participant per driving preference. Due to the limited representation of each driving preference in the validation and test sets (only one participant per category), the preference-specific results should be interpreted as held-out participant evaluations rather than generalizable preference population estimates. 

\begin{figure}[t]
	\centering
	\includegraphics[width=0.9\linewidth]{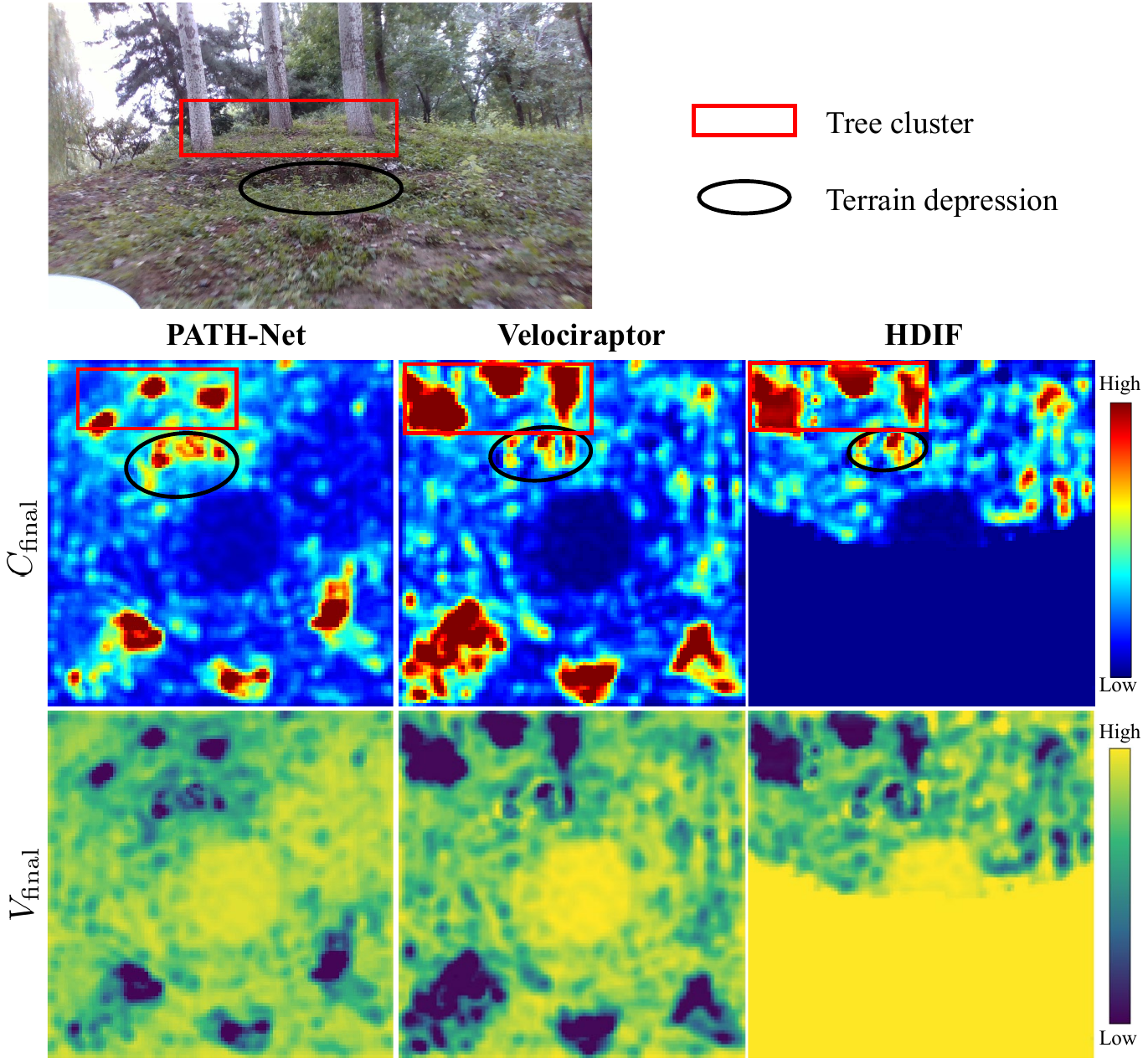}
    \caption{Qualitative comparison of cost-related and speed-related outputs for a representative held-out off-road test case. The second row shows the final costmaps, the third row shows the corresponding speed-related outputs. PATH-Net and Velociraptor generate spatial speedmaps, whereas HDIF generates a costmap using a fixed speed query, and its speed-related visualization is derived from the common planner rather than a learned speedmap.}
	\label{fig:qualitative_mhd}
\end{figure}

\subsubsection{Network Training}
Our PATH-Net, implemented in PyTorch, utilizes 5 parallel output heads ($N=5$). We trained the network for 50 epochs using Adam optimizer with a fixed learning rate ($1\times10^{-4}$). To ensure balanced representation across behavior categories (aggressive, neutral, and conservative), mini-batches are constructed with equal sampling from each category. During optimization, each output head is updated using an independent bootstrap sample drawn with replacement from its corresponding category-stratified subset, while the shared backbone is trained on the aggregated mean loss across all heads. This stratified sampling strategy promotes both general feature learning and task-specific adaptation, consistent with established methods in multi-objective learning \cite{tirinzoni2018importance, hassani2025towards}.

\subsubsection{MPPI Planner}
We use MPPI to estimate learner state-visitation frequencies for IRL and to generate recommended trajectories during inference. During inference, we evaluate the rollouts using the aggregated traversal cost, a penalty for exceeding the recommended speed bound, and the terminal goal distance.

\begin{table}[t]
	\centering
	\caption{Trajectory Alignment with Held-out Demonstrations (mean $\pm$ SD over Test Trajectories; lower is better).}
	\label{tab:mhd_results}
	\setlength{\tabcolsep}{3pt}
    \renewcommand{\arraystretch}{1.08}
    \resizebox{\columnwidth}{!}{%
    \begin{tabular}{llcc}
        \hline
        \textbf{Method}
        & \textbf{Metric}
        & \textbf{Aggressive}
        & \textbf{Conservative} \\
        \hline
        \multirow{2}{*}{Velociraptor}
        & pMHD (m)        & $0.37 \pm 0.08$ & $0.36 \pm 0.07$ \\
        & sMHD (m/s)      & $0.23 \pm 0.06$ & $0.21 \pm 0.05$ \\

        \multirow{2}{*}{HDIF}
        & pMHD (m)        & $0.39 \pm 0.09$ & $0.38 \pm 0.08$ \\
        & sMHD (m/s)      & --               & --               \\
        \hline
        \multirow{2}{*}{PATH-Net, neutral}
        & pMHD (m)        & $0.31 \pm 0.06$ & $0.29 \pm 0.06$ \\
        & sMHD (m/s)      & $0.18 \pm 0.04$ & $0.17 \pm 0.04$ \\

        \multirow{2}{*}{PATH-Net, aggressive}
        & pMHD (m)        & \textbf{0.17 $\pm$ 0.04} & $0.58 \pm 0.12$ \\
        & sMHD (m/s)      & \textbf{0.11 $\pm$ 0.03} & $0.38 \pm 0.08$ \\

        \multirow{2}{*}{PATH-Net, conservative}
        & pMHD (m)        & $0.61 \pm 0.13$ & \textbf{0.15 $\pm$ 0.03} \\
        & sMHD (m/s)      & $0.35 \pm 0.07$ & \textbf{0.10 $\pm$ 0.03} \\
        \hline
    \end{tabular}
    }
\end{table}

\subsection{Training Convergence and Validation Alignment}
\label{sec:convergence}
Fig.~\ref{fig:learning_curves}(a) shows the convergence characteristics of PATH-Net during training. Both the composite objective function $\mathcal{L}$ and its constituent terms --- the IRL loss $\mathcal{L}_{\mathrm{IRL}}$ and the speed regression loss $\mathcal{L}_{\mathrm{speed}}$ --- exhibit monotonic decrease during initial training epochs before reaching plateaus at approximately 20-30 epochs. This coordinated convergence behavior indicates effective joint optimization of the dual objectives under the proposed training paradigm.

A direct comparison of raw training losses across PATH-Net, HDIF, and Velociraptor is infeasible due to disparities in their learning objectives and loss scaling. To enable a fair evaluation, we employ the validation positional modified Hausdorff distance (pMHD) as a unified metric for trajectory alignment performance. As shown in Fig.~\ref{fig:learning_curves}(b), all models were trained for 50 epochs under identical conditions. The results demonstrate that PATH-Net achieves superior trajectory alignment, yielding a lower validation pMHD than both the adapted HDIF and Velociraptor baselines.

\subsection{Offline Trajectory Alignment}
\label{sec:mhd_analysis}
To quantify the fidelity of the generated recommended trajectories, we conduct an offline alignment analysis between predicted trajectories and held-out demonstrations. Trajectory alignment is evaluated using the modified Hausdorff distance (MHD) metric \cite{triest2024velociraptor}, which comprises two components: (i) positional accuracy (pMHD), computed as the bidirectional average nearest-neighbor Euclidean distance between position sets, and (ii) for speed-map prediction (sMHD), we use the bidirectional mean absolute speed difference in predicted versus demonstrated speed at spatially matched trajectory points.

We present a comparative evaluation of PATH-Net against two state-of-the-art traversability estimation methods: HDIF \cite{castro2023does} and Velociraptor (V-Dino) \cite{triest2024velociraptor}. The HDIF method estimates terrain-interaction costs based on visual-geometric observations and a fixed speed query~\cite{castro2023does} with $v_q=0.5\,\mathrm{m/s}$. Notably, while the costmap is queried using a fixed speed parameter, only the shared MPPI planner receives a constant speed bound for trajectory optimization and visualization. The V-Dino approach ($\nu=0.5$) predicts both cost and speed maps from visual-geometric features, though it does not account for operator-specific control preference \cite{triest2024velociraptor}. For fair comparison, both baselines were re-trained on our dataset and evaluated through the identical MPPI planner.

We focus on the aggressive and conservative held-out demonstrations because they represent the two most distinct preference settings in the dataset. The neutral setting is included as an intermediate PATH-Net configuration. Table~\ref{tab:mhd_results} reports the alignment results.

The aggressive preference setting of PATH-Net yields the lowest mean pMHD and sMHD for held-out aggressive trajectories, while the conservative preference setting yields the lowest values for held-out conservative trajectories. When these two preference settings are evaluated on trajectories from the other preference group, the pMHD and sMHD values become larger. This suggests that the aggregation parameter adjusts the recommended trajectories to align with the corresponding held-out participant's route choice and speed. The standard deviations in Table~\ref{tab:mhd_results} are computed across trajectories from a single held-out participant in each preference group, reflecting within-participant trajectory variation rather than between-participant uncertainty.

\paragraph{Qualitative Comparison}
Fig.~\ref{fig:qualitative_mhd} compares the cost-related and speed-related outputs of different approaches in a representative held-out test case. Both PATH-Net and Velociraptor generate spatial costmaps and speedmaps, HDIF generates a costmap queried at a fixed input speed, and does not predict a spatial speedmap. Therefore, the bottom row for HDIF shows the constant speed bound supplied to the common planner rather than a learned speedmap.

In this example, HDIF mainly reflects local terrain-interaction cost, whereas Velociraptor provides preference-independent cost and speed maps. PATH-Net produces localized high-cost regions while preserving a traversable corridor for the generated trajectory. This qualitative example supports the comparison in Table~\ref{tab:mhd_results}, but it does not separately evaluate the effects of the GFN, TCN, or aggregation modules.

\begin{figure}[t]
	\centering
\includegraphics[width=0.8\linewidth]{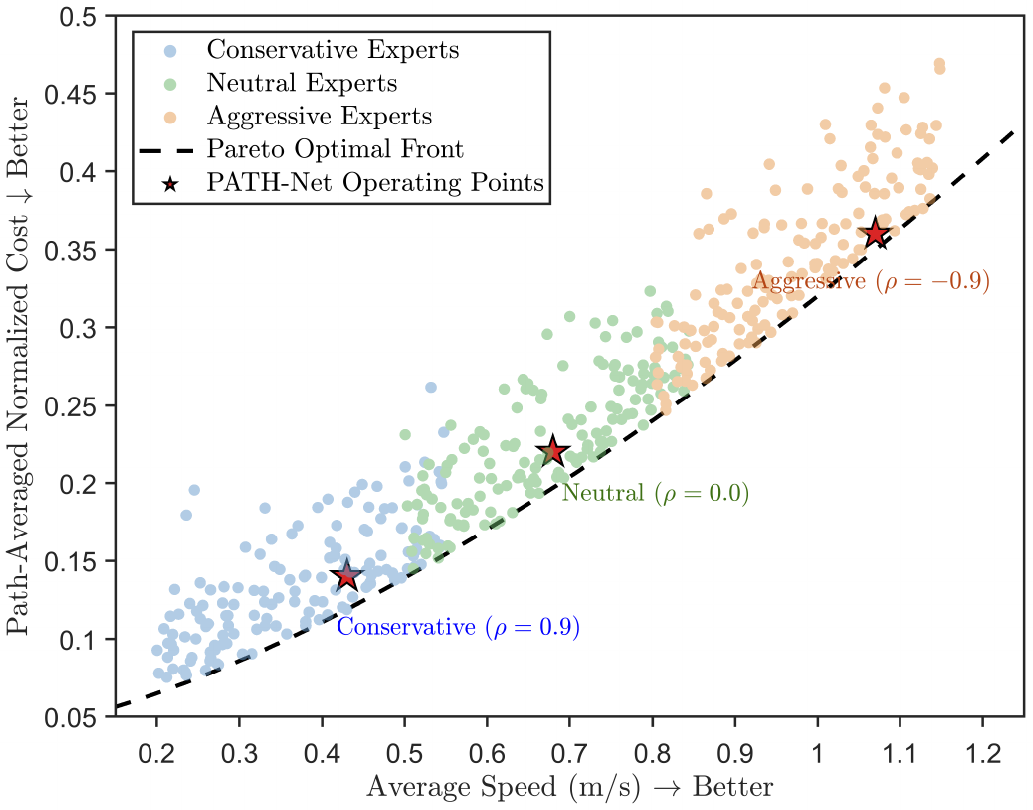} 
    \caption{Dataset-level operating points comparison. Human demonstrations are shown by their executed speed and path-averaged cost. Each PATH-Net marker shows the mean recommended speed and mean path-averaged cost obtained by applying one preference setting to the same recorded cases. The curve indicates the non-dominated frontier of the human demonstrations.}
	\label{fig:policy_frontier}
\end{figure}

\subsection{Questionnaire-Based Workload Assessment}
\label{sec:workload_assessment}
We used a within-subject questionnaire to compare self-reported expected workload with and without PATH-Net recommendations. In the first stage, participants evaluated the expected workload of teleoperation without additional traversability information. In the second stage, they evaluated the same type of task after reviewing PATH-Net route and speed recommendations.

Following each experimental trial, participants evaluated their subjective workload using the NASA-TLX questionnaire \cite{noyes2007self}. To maintain conceptual consistency across all workload dimensions, the performance item was reverse-scored according to
\begin{equation} 
S_{\mathrm{burden}} = 100-S_{\mathrm{ success}}. 
\label{eq:performance_burden} 
\end{equation}
where $S_{\mathrm{success}}$ denotes the raw success score. This transformation aligns all six NASA-TLX subscales so that higher values uniformly indicate greater reported workload.

As shown in Table~\ref{tab:nasa_tlx}, participants reported a lower mean NASA-TLX overall workload score with PATH-Net recommendations than without recommendations ($22.2$ versus $34.7$). The relative decrease was $36.0\%$. Because the two information conditions were presented in a fixed order, this result should be interpreted descriptively rather than as causal evidence of workload reduction.

\begin{table}[t]
	\centering
	\caption{NASA-TLX Workload Ratings under Different Recommendation Conditions. SD values are shown in gray parentheses.}
	\label{tab:nasa_tlx}
	\footnotesize
	\setlength{\tabcolsep}{3pt}
	\resizebox{\columnwidth}{!}{%
	\begin{tabular}{lccc}
		\hline
		\textbf{Workload Subscale} & \textbf{No Recommendation} & \textbf{With PATH-Net} & \textbf{Decrease (\%)} \\ \hline
		Mental Demand       & \msd{42.3}{21.4} & \msd{31.3}{17.5} & 26.0 \\
        Physical Demand     & \msd{24.1}{17.9} & \msd{15.6}{13.4} & 35.2 \\
        Temporal Demand     & \msd{34.3}{20.4} & \msd{19.5}{11.7} & 43.2 \\
        Performance Burden  & \msd{30.1}{22.7} & \msd{25.1}{22.5} & 16.6 \\
        Effort              & \msd{46.4}{20.5} & \msd{30.1}{18.2} & 35.2 \\
        Frustration         & \msd{31.0}{15.0} & \msd{11.6}{12.3} & 62.6 \\
        \hline
		Overall Average     & \msd{34.7}{14.0} & \msd{22.2}{10.0} & 36.0 \\
        \hline
	\end{tabular}
	}
\end{table}

\subsection{Preference Effects on Recommended Trajectories}
\label{sec:personalization_analysis}
We examine how the preference parameter changes the generated route and speed profile.

\begin{table}[t]
	\centering
    \caption{Recommended Trajectory Measures under Different Preference Settings. Values are averaged over held-out test cases.}
	\label{tab:style_path_statistics}
	\footnotesize
	\setlength{\tabcolsep}{2pt}
	\resizebox{\columnwidth}{!}{%
	\begin{tabular}{lccc}
		\hline
		\textbf{Preference Setting ($\rho$)} & \textbf{Min. Clearance (m)} & \textbf{Avg. Recommended Speed (m/s)} & \textbf{Path-Averaged Cost} \\ \hline
		Aggressive ($-0.9$) & $0.18$ & $1.07$ & $0.36$ \\
		Neutral ($0.0$) & $0.35$ & $0.68$ & $0.22$ \\
		Conservative ($+0.9$) & $0.62$ & $0.43$ & $0.14$ \\ \hline
	\end{tabular}
	}
\end{table}

\begin{figure}[t]
	\centering
	\includegraphics[width=0.8\linewidth]{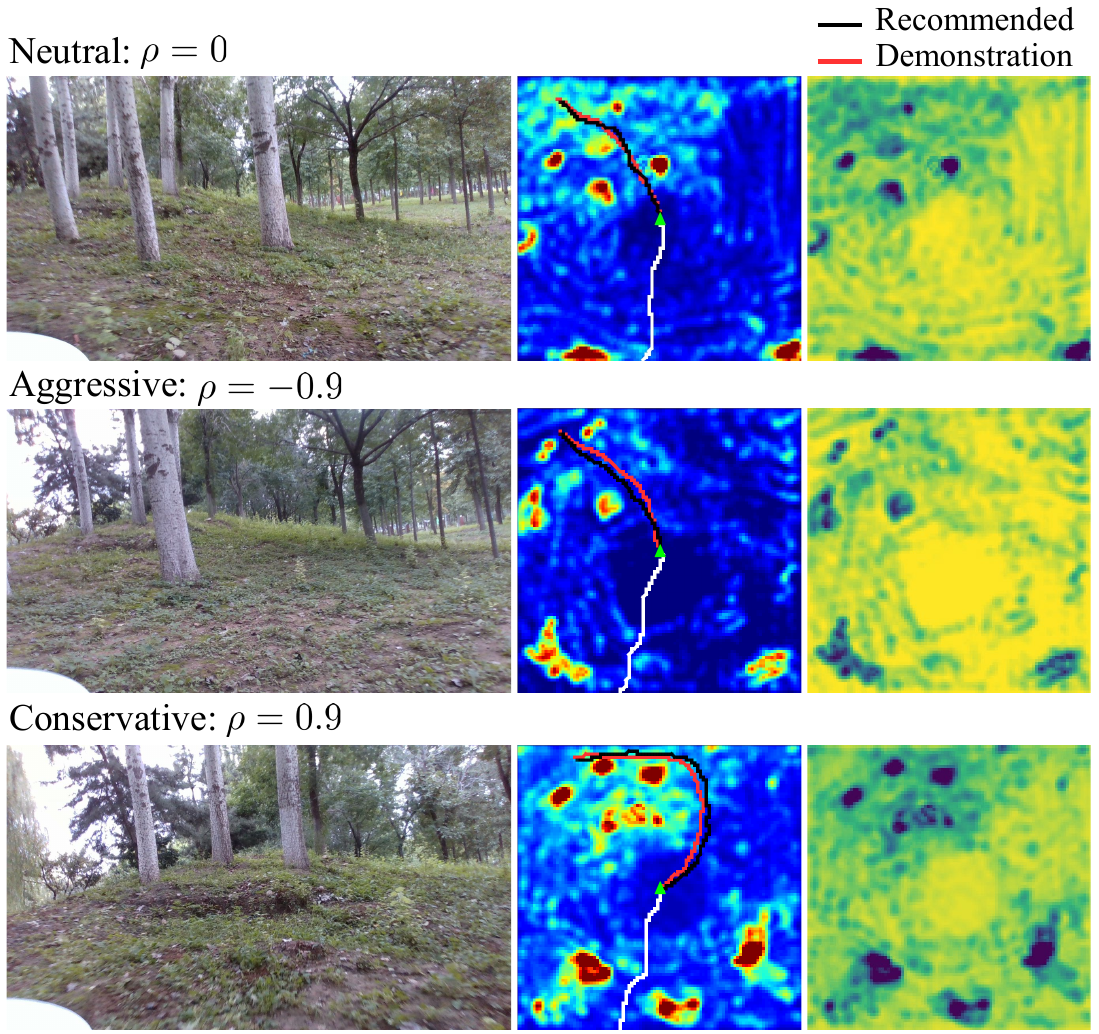}
	\caption{Examples of PATH-Net recommended trajectories under different preference settings with the historic trajectory (white curve), held-out demonstration (red curve), and PATH-Net recommended trajectory (black curve).}
	\label{fig:path_cvar}
\end{figure}

\begin{figure*}[t]
	\centering
	\includegraphics[width=0.9\linewidth]{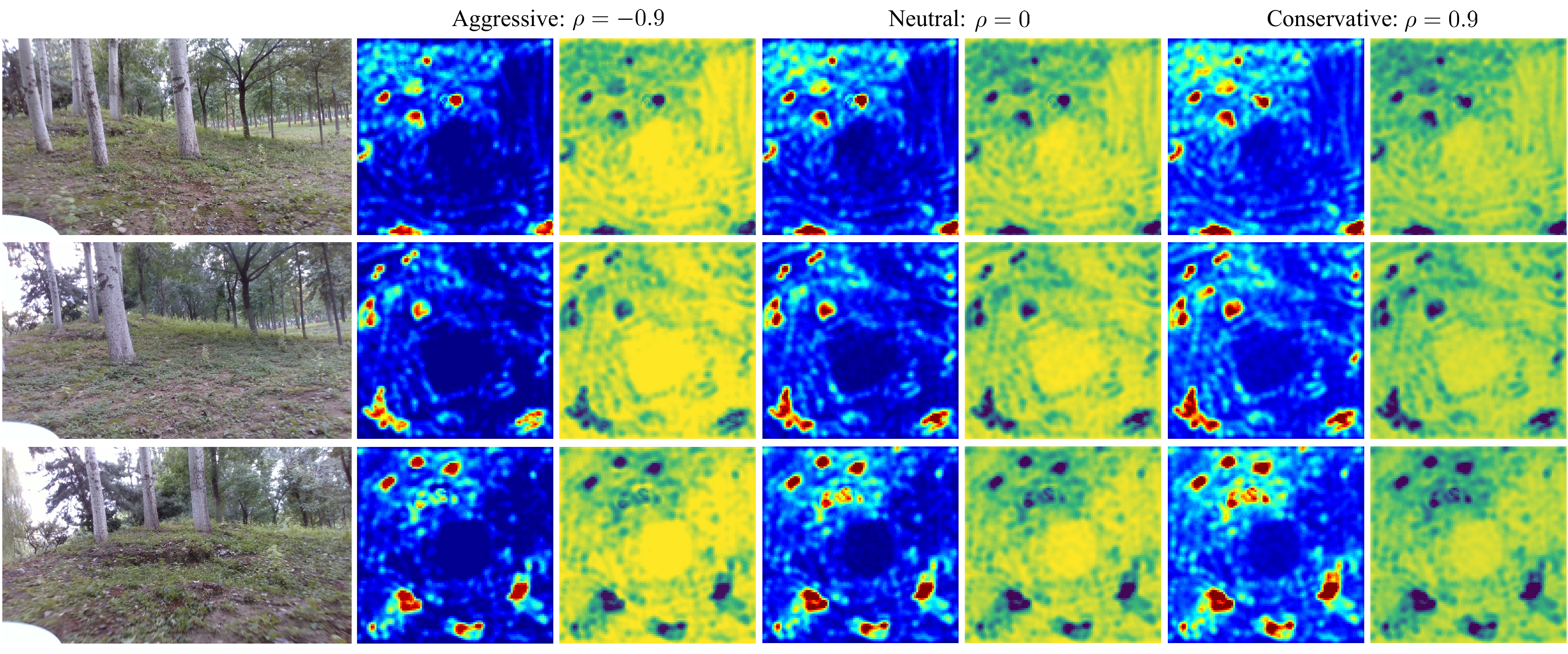}
    \caption{Aggregated costmaps and speedmaps under different preference settings. All costmaps use the range $[0,1]$, and all speedmaps share the range $[0,v_{\max}]$. Moving from the aggressive to the conservative setting increases the contribution of larger candidate costs and lowers the recommended speed bound.}
	\label{fig:qualitative_style}
\end{figure*}

\paragraph{Recommended Operating-Point Analysis}
Table~\ref{tab:style_path_statistics} reports recommended trajectory measures on the held-out test cases. We also use all collected demonstrations to compare the generated trajectories with the overall distribution of operator behavior. To compare trajectories on a common cost scale, we define the preference-independent ensemble-mean costmap as
\begin{equation} 
\overline{C}(\mathbf{g}) = \frac{1}{N} \sum_{i=1}^{N} C_i(\mathbf{g}).
\label{eq:base_costmap} 
\end{equation}
For a trajectory $\tau=\{(\mathbf{g}_q,v_q)\}_{q=1}^{L_{\tau}}$, the path-averaged cost is
\begin{equation}
R(\tau) = \frac{1}{L_{\tau}} \sum_{q=1}^{L_{\tau}} \overline{C}(\mathbf{g}_q), 
\label{eq:path_cost}
\end{equation}
and the average speed is
\begin{equation}
E(\tau) = \frac{1}{L_{\tau}} \sum_{q=1}^{L_{\tau}} v_q. 
\label{eq:average_speed} 
\end{equation}
For human demonstrations, $E(\tau)$ denotes the executed speed; for PATH-Net trajectories, it denotes the recommended speed generated by the planner.

A demonstration is Pareto-efficient if no other demonstration has both a lower path-averaged cost and a higher average speed, with at least one strict improvement. The non-dominated frontier in Fig.~\ref{fig:policy_frontier} is computed from all collected human demonstrations. For each preference parameter $\rho$, PATH-Net is applied to the same $M$ recorded cases. The corresponding star denotes the mean operating point
\begin{equation}
\left( \overline{E}_{\rho}, \overline{R}_{\rho} \right) = \left( \frac{1}{M} \sum_{m=1}^{M} E\!\left(\tau_{\rho,m}^{*}\right), \frac{1}{M} \sum_{m=1}^{M} R\!\left(\tau_{\rho,m}^{*}\right) \right). 
\label{eq:mean_style_operating_point} 
\end{equation}
The aggressive setting produces trajectories with higher average speed and higher path-averaged cost, whereas the conservative setting produces trajectories with lower average speed and lower path-averaged cost.

\paragraph{Qualitative Preference Comparison}
Fig.~\ref{fig:qualitative_style} illustrates how the aggregated costmap and speedmap vary with $\rho$. All costmaps use the same color range $[0,1]$, and all speedmaps use the same range $[0,v_{\max}]$. As $\rho$ increases, the upper-tail cost aggregation places greater weight on larger candidate costs, while the speed quantile selects a lower candidate speed. In the displayed cases, this produces larger high-cost regions around some obstacles and lower speed bounds in the corresponding areas.

Fig.~\ref{fig:path_cvar} shows representative recommended trajectories under the three preference settings. The white curve denotes the previously traversed path, the red curve denotes the held-out demonstration, and the black curve denotes the PATH-Net recommended trajectory. In the aggressive example, the recommended trajectory passes closer to the terrain boundary and is visually similar to the corresponding aggressive demonstration. In the conservative example, upper-tail cost aggregation increases the influence of larger candidate costs, and the resulting recommended trajectory maintains greater clearance.

The panels show different held-out test cases. They therefore illustrate how PATH-Net outputs vary across preference settings, but do not isolate the effect of $\rho$ under identical environmental and vehicle-state inputs.

\section{Conclusion and Future Work}
\label{sec:conclusion}
This paper presented PATH-Net, a decision-support framework for generating preference-conditioned cost and speed maps in off-road teleoperation. PATH-Net combines visual--geometric BEV perception with recent vehicle-motion history and learns multiple candidate cost and speed maps from trained-operator demonstrations. An asymmetric aggregation mechanism applies upper-tail aggregation to the candidate cost maps and quantile selection to the candidate speed maps, allowing the recommended outputs to vary with a questionnaire-derived preference parameter.

Under the participant-disjoint evaluation, the aggressive and conservative settings achieved the lowest pMHD and sMHD values for the corresponding held-out demonstrations. The preference analysis showed that the conservative setting produced lower recommended speed, lower path-averaged cost, and larger obstacle clearance than the aggressive setting. In the exploratory questionnaire assessment, participants reported a lower mean NASA-TLX-based workload score with PATH-Net recommendations than without them ($22.2$ versus $34.7$). Because the two information conditions were presented in a fixed order, this difference should be interpreted descriptively rather than as causal evidence of workload reduction.

The present evaluation is limited by the number of participants, including one held-out participant per preference category in the validation and test sets, and by the fixed-order questionnaire design. Future work will examine online estimation of operator preferences from recent control behavior, incorporate additional vehicle--terrain interaction measurements such as wheel slip and suspension response, and evaluate the framework with larger participant samples and counterbalanced human-subject protocols. These results support further investigation of preference-conditioned trajectory and speed recommendations for teleoperated off-road navigation.
		%\section*{Acknowledgment}The authors would like to thank...

		\bibliographystyle{IEEEtran}
		\bibliography{refs}

@article{dinelli2023configurations,
  title={Configurations and applications of multi-agent hybrid drone/unmanned ground vehicle for underground environments: A review},
  author={Dinelli, Chris and Racette, John and Escarcega, Mario and Lotero, Simon and Gordon, Jeffrey and Montoya, James and Dunaway, Chase and Androulakis, Vasileios and Khaniani, Hassan and Shao, Sihua and others},
  journal={Drones},
  volume={7},
  number={2},
  pages={136},
  year={2023},
  publisher={MDPI}
}

@article{li2023classification,
  title={The classification and new trends of shared control strategies in telerobotic systems: A survey},
  author={Li, Gaofeng and Li, Qiang and Yang, Chenguang and Su, Yuan and Yuan, Zuqiang and Wu, Xinyu},
  journal={IEEE Transactions on Haptics},
  volume={16},
  number={2},
  pages={118--133},
  year={2023},
  publisher={IEEE}
}

@article{du2024sensory,
  title={Sensory manipulation as a countermeasure to robot teleoperation delays: system and evidence},
  author={Du, Jing and Vann, William and Zhou, Tianyu and Ye, Yang and Zhu, Qi},
  journal={Scientific Reports},
  volume={14},
  number={1},
  pages={4333},
  year={2024},
  publisher={Nature Publishing Group UK London}
}

@article{benrabah2024review,
  title={A review on traversability risk assessments for autonomous ground vehicles: Methods and metrics},
  author={Benrabah, Mohamed and Orou Mousse, Charifou and Randriamiarintsoa, Elie and Chapuis, Roland and Aufr{\`e}re, Romuald},
  journal={Sensors},
  volume={24},
  number={6},
  pages={1909},
  year={2024},
  publisher={MDPI}
}

@article{asvadi20163d,
  title={3D Lidar-based static and moving obstacle detection in driving environments: An approach based on voxels and multi-region ground planes},
  author={Asvadi, Alireza and Premebida, Cristiano and Peixoto, Paulo and Nunes, Urbano},
  journal={Robotics and Autonomous Systems},
  volume={83},
  pages={299--311},
  year={2016},
  publisher={Elsevier}
}

@article{e2025terrain,
  title={On terrain traversability analysis in unstructured environments: recent advances in forest applications},
  author={E. Carvalho, Afonso and Portugal, David and Peixoto, Paulo},
  journal={Intelligent Service Robotics},
  volume={18},
  number={2},
  pages={195--213},
  year={2025},
  publisher={Springer}
}

@article{bhosale2026review,
  title={A Review of Learning Off-Road Terrain Traversability for Autonomous Ground Vehicles},
  author={Bhosale, Mayuresh and Whitson, Jordan A and Jia, Yunyi},
  journal={Journal of Autonomous Vehicles and Systems},
  pages={1--24},
  year={2026}
}

@inproceedings{triest2024velociraptor,
  title={Velociraptor: Leveraging visual foundation models for label-free, risk-aware off-road navigation},
  author={Triest, Samuel and Sivaprakasam, Matthew and Aich, Shubhra and Fan, David and Wang, Wenshan and Scherer, Sebastian},
  booktitle={8th Annual Conference on Robot Learning},
  year={2024}
}

@inproceedings{castro2023does,
  title={How does it feel? self-supervised costmap learning for off-road vehicle traversability},
  author={Castro, Mateo Guaman and Triest, Samuel and Wang, Wenshan and Gregory, Jason M and Sanchez, Felix and Rogers, John G and Scherer, Sebastian},
  booktitle={2023 IEEE International Conference on Robotics and Automation (ICRA)},
  pages={931--938},
  year={2023},
  organization={IEEE}
}

@article{oquab2023dinov2,
  title={Dinov2: Learning robust visual features without supervision},
  author={Oquab, Maxime and Darcet, Timoth{\'e}e and Moutakanni, Th{\'e}o and Vo, Huy and Szafraniec, Marc and Khalidov, Vasil and Fernandez, Pierre and Haziza, Daniel and Massa, Francisco and El-Nouby, Alaaeldin and others},
  journal={arXiv preprint arXiv:2304.07193},
  year={2023}
}

@article{englert2025vfm,
  title={VFM-UDA++: Improving Network Architectures and Data Strategies for Unsupervised Domain Adaptive Semantic Segmentation},
  author={Englert, Brun{\'o} B and Dubbelman, Gijs},
  journal={arXiv preprint arXiv:2503.10685},
  year={2025}
}

@inproceedings{sivaprakasam2025salon,
  title={Salon: Self-supervised adaptive learning for off-road navigation},
  author={Sivaprakasam, Matthew and Triest, Samuel and Ho, Cherie and Aich, Shubhra and Lew, Jeric and Adu, Isaiah and Wang, Wenshan and Scherer, Sebastian},
  booktitle={2025 IEEE International Conference on Robotics and Automation (ICRA)},
  pages={16999--17006},
  year={2025},
  organization={IEEE}
}

@inproceedings{jung2024v,
  title={V-strong: Visual self-supervised traversability learning for off-road navigation},
  author={Jung, Sanghun and Lee, JoonHo and Meng, Xiangyun and Boots, Byron and Lambert, Alexander},
  booktitle={2024 IEEE International Conference on Robotics and Automation (ICRA)},
  pages={1766--1773},
  year={2024},
  organization={IEEE}
}

@article{bae2023self,
  title={Self-supervised 3D traversability estimation with proxy bank guidance},
  author={Bae, Jihwan and Seo, Junwon and Kim, Taekyung and Jeon, Hae-Gon and Kwak, Kiho and Shim, Inwook},
  journal={IEEE Access},
  volume={11},
  pages={51490--51501},
  year={2023},
  publisher={IEEE}
}

@article{wang2024traversability,
  title={Traversability estimation for off-road autonomous driving under ego-motion uncertainty},
  author={Wang, Kai and Wang, Meiling and Wang, Rongchuan and Zhai, Chaoyang and Song, Wenjie},
  journal={IEEE Sensors Journal},
  volume={24},
  number={5},
  pages={6584--6596},
  year={2024},
  publisher={IEEE}
}

@article{zhang2024autonomous,
  title={Autonomous vehicles traversability mapping fusing semantic--geometric in off-road navigation},
  author={Zhang, Bo and Chen, Weili and Xu, Chaoming and Qiu, Jinshi and Chen, Shiyu},
  journal={Drones},
  volume={8},
  number={9},
  pages={496},
  year={2024},
  publisher={MDPI}
}

@article{frey2024roadrunner,
  title={RoadRunner—Learning traversability estimation for autonomous off-road driving},
  author={Frey, Jonas and Patel, Manthan and Atha, Deegan and Nubert, Julian and Fan, David and Agha, Ali and Padgett, Curtis and Spieler, Patrick and Hutter, Marco and Khattak, Shehryar},
  journal={IEEE Transactions on Field Robotics},
  volume={1},
  pages={192--212},
  year={2024},
  publisher={IEEE}
}

@inproceedings{jeon2024follow,
  title={Follow the footprints: Self-supervised traversability estimation for off-road vehicle navigation based on geometric and visual cues},
  author={Jeon, Yurim and Son, E In and Seo, Seung-Woo},
  booktitle={2024 IEEE International Conference on Robotics and Automation (ICRA)},
  pages={1774--1780},
  year={2024},
  organization={IEEE}
}

@inproceedings{liang2025gnd,
  title={Gnd: Global navigation dataset with multi-modal perception and multi-category traversability in outdoor campus environments},
  author={Liang, Jing and Das, Dibyendu and Song, Daeun and Shuvo, Md Nahid Hasan and Durrani, Mohammad and Taranath, Karthik and Penskiy, Ivan and Manocha, Dinesh and Xiao, Xuesu},
  booktitle={2025 IEEE International Conference on Robotics and Automation (ICRA)},
  pages={2383--2390},
  year={2025},
  organization={IEEE}
}

@inproceedings{paton2020navigation,
  title={Navigation on the line: Traversability analysis and path planning for extreme-terrain rappelling rovers},
  author={Paton, Michael and Strub, Marlin P and Brown, Travis and Greene, Rebecca J and Lizewski, Jacob and Patel, Vandan and Gammell, Jonathan D and Nesnas, Issa AD},
  booktitle={2020 IEEE/RSJ International Conference on Intelligent Robots and Systems (IROS)},
  pages={7034--7041},
  year={2020},
  organization={IEEE}
}

@inproceedings{leininger2024gaussian,
  title={Gaussian process-based traversability analysis for terrain mapless navigation},
  author={Leininger, Abe and Ali, Mahmoud and Jardali, Hassan and Liu, Lantao},
  booktitle={2024 IEEE International Conference on Robotics and Automation (ICRA)},
  pages={10925--10931},
  year={2024},
  organization={IEEE}
}

@article{triest2023learning,
  title={Learning risk-aware costmaps via inverse reinforcement learning for off-road navigation},
  author={Triest, Samuel and Castro, Mateo Guaman and Maheshwari, Parv and Sivaprakasam, Matthew and Wang, Wenshan and Scherer, Sebastian},
  journal={arXiv preprint arXiv:2302.00134},
  year={2023}
}

@article{fan2021learning,
  title={Learning risk-aware costmaps for traversability in challenging environments},
  author={Fan, David D and Agha-Mohammadi, Ali-Akbar and Theodorou, Evangelos A},
  journal={IEEE robotics and automation letters},
  volume={7},
  number={1},
  pages={279},
  year={2021}
}

@article{dixit2024step,
  title={Step: Stochastic traversability evaluation and planning for risk-aware navigation; results from the darpa subterranean challenge},
  author={Dixit, Anushri and Fan, David D and Otsu, Kyohei and Dey, Sharmita and Agha-Mohammadi, Ali-Akbar and Burdick, Joel},
  journal={Field Robotics},
  volume={4},
  pages={182--210},
  year={2024},
  publisher={FRPS}
}

@article{borges2022survey,
  title={A survey on terrain traversability analysis for autonomous ground vehicles: Methods, sensors, and challenges},
  author={Borges, Paulo VK and Peynot, Thierry and Liang, Sisi and Arain, Bilal and Wildie, Matthew and Minareci, Melih G and Lichman, Serge and Samvedi, Garima and Sa, Inkyu and Hudson, Nicolas and others},
  journal={Field Robotics},
  volume={2},
  pages={1567--1627},
  year={2022},
  publisher={FRPS}
}

@inproceedings{fusaro2026learning,
   title={Pushing the Limits of Learning-Based Traversability Analysis for Autonomous Driving on CPU},
   author={Fusaro, Daniel and Olivastri, Emilio and Evangelista, Daniele and Imperoli, Marco and Menegatti, Emanuele and Pretto, Alberto},
   booktitle={Proceedings of the 17th International Conference on Intelligent Autonomous Systems},
   pages={457--469},
   year={2022},
   publisher={Springer},
   doi={10.1007/978-3-031-22216-0_36}
}

@article{zhang2018integrating,
  title={Integrating kinematics and environment context into deep inverse reinforcement learning for predicting off-road vehicle trajectories},
  author={Zhang, Yanfu and Wang, Wenshan and Bonatti, Rogerio and Maturana, Daniel and Scherer, Sebastian},
  journal={arXiv preprint arXiv:1810.07225},
  year={2018}
}

@article{fankhauser2018probabilistic,
  title={Probabilistic terrain mapping for mobile robots with uncertain localization},
  author={Fankhauser, P{\'e}ter and Bloesch, Michael and Hutter, Marco},
  journal={IEEE Robotics and Automation Letters},
  volume={3},
  number={4},
  pages={3019--3026},
  year={2018},
  publisher={IEEE}
}

@article{sevastopoulos2022survey,
  title={A survey of traversability estimation for mobile robots},
  author={Sevastopoulos, Christos and Konstantopoulos, Stasinos},
  journal={IEEE Access},
  volume={10},
  pages={96331--96347},
  year={2022},
  publisher={IEEE}
}

@article{lin2021long,
  title={Long-horizon humanoid navigation planning using traversability estimates and previous experience},
  author={Lin, Yu-Chi and Berenson, Dmitry},
  journal={Autonomous Robots},
  volume={45},
  number={6},
  pages={937--956},
  year={2021},
  publisher={Springer}
}

@article{cladera2025air,
  title={Air-ground collaboration for language-specified missions in unknown environments},
  author={Cladera, Fernando and Ravichandran, Zachary and Hughes, Jason and Murali, Varun and Nieto-Granda, Carlos and Hsieh, M Ani and Pappas, George J and Taylor, Camillo J and Kumar, Vijay},
  journal={IEEE Transactions on Field Robotics},
  year={2025},
  publisher={IEEE}
}

@article{shen2024towards,
  title={Towards bidirectional human-ai alignment: A systematic review for clarifications, framework, and future directions},
  author={Shen, Hua and Knearem, Tiffany and Ghosh, Reshmi and Alkiek, Kenan and Krishna, Kundan and Liu, Yachuan and Ma, Ziqiao and Petridis, Savvas and Peng, Yi-Hao and Qiwei, Li and others},
  journal={arXiv preprint arXiv:2406.09264},
  volume={2406},
  pages={1--56},
  year={2024}
}

@article{hoc2009cooperation,
  title={Cooperation between drivers and automation: implications for safety},
  author={Hoc, Jean-Michel and Young, Mark S and Blosseville, Jean-Marc},
  journal={Theoretical Issues in Ergonomics Science},
  volume={10},
  number={2},
  pages={135--160},
  year={2009},
  publisher={Taylor \& Francis}
}

@article{nan2025safe,
  title={Safe and interpretable human-like planning with transformer-based deep inverse reinforcement learning for autonomous driving},
  author={Nan, Jiangfeng and Zhang, Ruzheng and Yin, Guodong and Zhuang, Weichao and Zhang, Yilong and Deng, Weiwen},
  journal={IEEE Transactions on Automation Science and Engineering},
  volume={22},
  pages={12134--12146},
  year={2025},
  publisher={IEEE}
}

@article{hong2014monte,
  title={Monte Carlo methods for value-at-risk and conditional value-at-risk: a review},
  author={Hong, L Jeff and Hu, Zhaolin and Liu, Guangwu},
  journal={ACM Transactions on Modeling and Computer Simulation (TOMACS)},
  volume={24},
  number={4},
  pages={1--37},
  year={2014},
  publisher={ACM New York, NY, USA}
}

@article{zhang2022guided,
  title={Guided filter network for semantic image segmentation},
  author={Zhang, Xiang and Zhao, Wanqing and Zhang, Wei and Peng, Jinye and Fan, Jianping},
  journal={IEEE Transactions on Image Processing},
  volume={31},
  pages={2695--2709},
  year={2022},
  publisher={IEEE}
}

@article{jia2021residual,
  title={Residual-guided in-loop filter using convolution neural network},
  author={Jia, Wei and Li, Li and Li, Zhu and Zhang, Xiang and Liu, Shan},
  journal={ACM Transactions on Multimedia Computing, Communications, and Applications (TOMM)},
  volume={17},
  number={4},
  pages={1--19},
  year={2021},
  publisher={ACM New York, NY}
}

@inproceedings{tirinzoni2018importance,
  title={Importance weighted transfer of samples in reinforcement learning},
  author={Tirinzoni, Andrea and Sessa, Andrea and Pirotta, Matteo and Restelli, Marcello},
  booktitle={International Conference on Machine Learning},
  pages={4936--4945},
  year={2018},
  organization={PMLR}
}

@article{hassani2025towards,
  title={Towards sample-efficiency and generalization of transfer and inverse reinforcement learning: A comprehensive literature review},
  author={Hassani, Hossein and Hallaji, Ehsan and Razavi-Far, Roozbeh and Saif, Mehrdad and Lin, Liang},
  journal={IEEE Transactions on Artificial Intelligence},
  year={2025},
  publisher={IEEE}
}

@article{rolison2023psychological,
  title={The psychological drivers of self-reported risk preference.},
  author={Rolison, Jonathan J},
  journal={Journal of experimental psychology: learning, memory, and cognition},
  volume={49},
  number={10},
  pages={1635},
  year={2023},
  publisher={American Psychological Association}
}

@inproceedings{matsushita2021operator,
  title={An Operator and Observer Style Teleoperation User Interface for Remote Object Search},
  author={Matsushita, Masanori and Hayashi, Kotaro and Miura, Jun},
  booktitle={2021 IEEE/SICE International Symposium on System Integration (SII)},
  pages={762--768},
  year={2021},
  organization={IEEE}
}

@article{noyes2007self,
  title={A self-analysis of the NASA-TLX workload measure},
  author={Noyes, Jan M and Bruneau, Daniel PJ},
  journal={Ergonomics},
  volume={50},
  number={4},
  pages={514--519},
  year={2007},
  publisher={Taylor \& Francis}
}

@inproceedings{quigley2009ros,
  title={ROS: an open-source Robot Operating System},
  author={Quigley, Morgan and Conley, Ken and Gerkey, Brian and Faust, Josh and Foote, Tully and Leibs, Jeremy and Wheeler, Rob and Ng, Andrew Y and others},
  booktitle={ICRA workshop on open source software},
  volume={3},
  number={3.2},
  pages={5},
  year={2009},
  organization={Kobe}
}
		
		\ifCLASSOPTIONcaptionsoff
		\newpage
		\fi
		\begin{IEEEbiography}[{\includegraphics[width=1in,height=1.25in,clip,keepaspectratio]{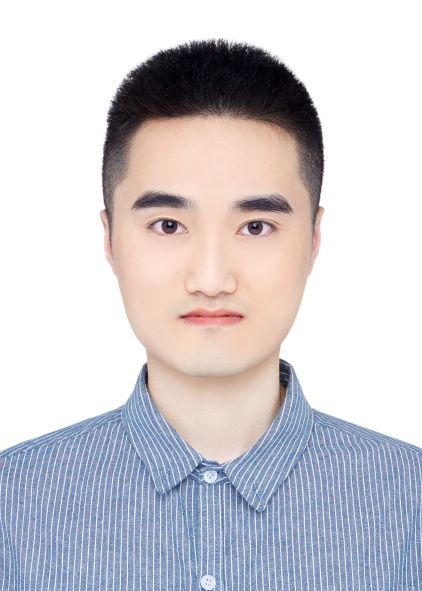}}]{Xianghao Meng} received the B.S. degree in mechanical engineering from Beijing University of Chemical Technology, China, in 2021. He is currently pursuing the Ph.D. degree in Beijing Institute of Technology, China. His research interests include intelligent vehicles, human-machine collaborative trust evaluation, calibration, and modeling, agent capability evaluation, and human-machine transparency.
		\end{IEEEbiography}
		
		\begin{IEEEbiography}[{\includegraphics[width=1in,height=1.25in,clip,keepaspectratio]{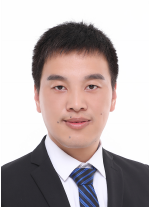}}]{Wenshuo Wang} (SM'15-M'18) received his Ph.D. degree in mechanical engineering from the Beijing Institute of Technology (BIT) in 2018.  Presently, he is a Full Professor at the School of Mechanical Engineering, BIT, Beijing, China. Prior to his role at BIT, he completed Postdoctoral fellowships at McGill University, Carnegie Mellon University (CMU), and UC Berkeley between 2018 and 2023. Furthermore, from 2015 to 2018, he served as a Research Assistant at UC Berkeley and the University of Michigan, Ann Arbor. His research interests focus on Bayesian nonparametric learning, human driver model, human–vehicle interaction, ADAS, and autonomous vehicles.
		\end{IEEEbiography}
		
		\begin{IEEEbiography} [{\includegraphics[width=1in,height=1.25in,clip,keepaspectratio]{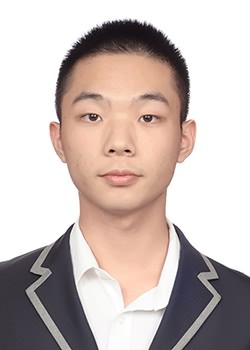}}]{Cheng Shao} received the B.S. degree in mechanical engineering from Beijing Institute of Technology, China, in 2024. He is currently pursuing the Master's degree in Beijing Institute of Technology, China, His research interests include intelligent vehicles, Human Intention Recognition, shared autonomy, and Human-Vehicle trust.
		\end{IEEEbiography}

        \begin{IEEEbiography}[{\includegraphics[width=1in,height=1.25in,clip,keepaspectratio]{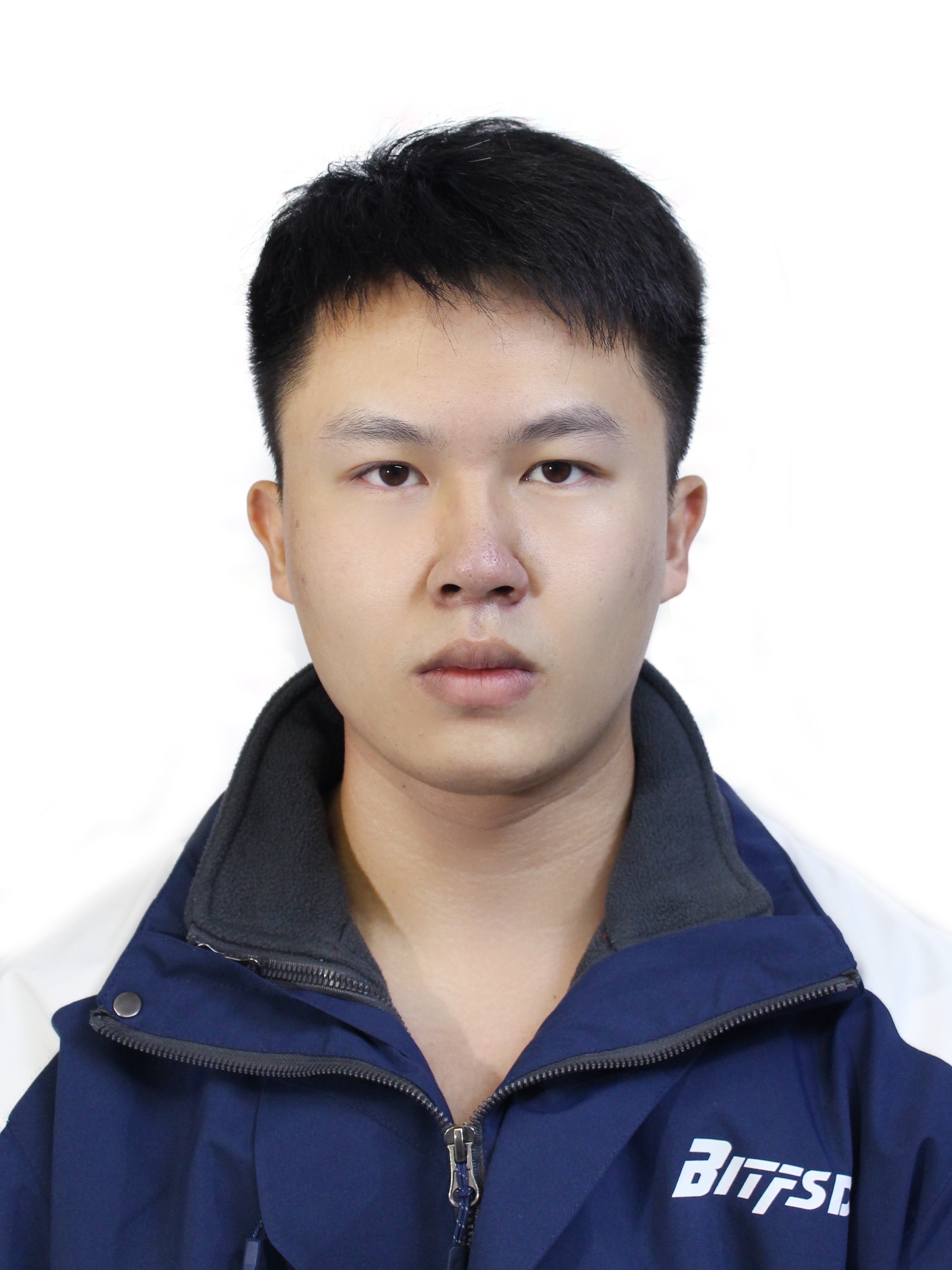}}]{Lewei Feng} received his B.S. degree in vehicle engineering from the Beijing Institute of Technology (BIT), Beijing, China, in 2023, where he is currently pursuing the Ph.D. degree with the Intelligent Vehicle Research Center, School of Mechanical Engineering, Beijing Institute of Technology. His research interests include teleoperation in unstructured environments,  traversability assessment and motion planning.
        \end{IEEEbiography}
		
		% \begin{IEEEbiography} [{\includegraphics[width=1in,height=1.25in,clip,keepaspectratio]{Figures/HuiYao}}]{Hui Yao} received the B.S. degree in mechanical engineering from Beijing Institute of Technology (BIT), Beijing, China, in 2018. She is currently working toward the Ph.D. degree in mechanical engineering from Beijing Institute of Technology (BIT), Beijing, China. Her research interests include driver intention detection, machine learning, brain-machine intelligence and human-intelligence vehicle collaboration.
		% \end{IEEEbiography}
		
		% \begin{IEEEbiography} [{\includegraphics[width=1in,height=1.25in,clip,keepaspectratio]{Figures/RuizengZhang}}]{Ruizeng Zhang} received the B.E. degree from Shandong University, Jinan, China, in 2016. He is currently a postdoctoral fellow with the School of Mechanical Engineering, Beijing Institute of Technology. He was a Visiting Scholar with the Mechatronic Vehicle Systems Lab, University of Waterloo from 2021 to 2023. His main research interest are motion control and planning of autonomous vehicle system.
		% \end{IEEEbiography}

        \begin{IEEEbiography}[{\includegraphics[width=1in,height=1.25in,clip,keepaspectratio]{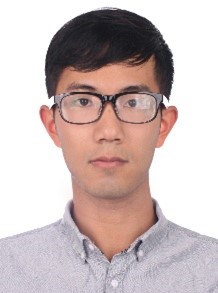}}]{Qingxiao Liu} received the B.S. and Ph.D. degrees in mechanical engineering from Beijing Institute of Technology, Beijing, China, in 2018 and 2025, respectively. He is currently a Postdoctoral Fellow in the Department of Industrial and Systems Engineering, The Hong Kong Polytechnic University, Hong Kong, China. His research interests include intelligent vehicle environment perception and understanding, decision-making, path/motion planning, and control.
        \end{IEEEbiography}

        \begin{IEEEbiography}[{\includegraphics[width=1in,height=1.25in,clip,keepaspectratio]{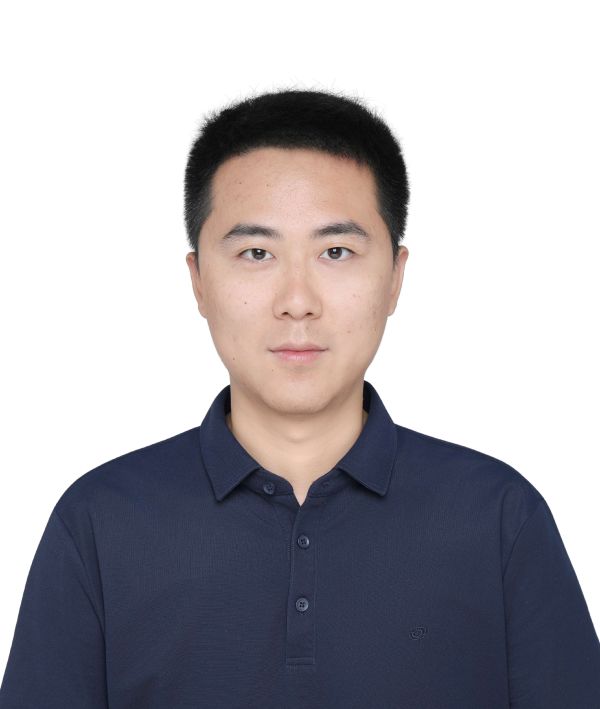}}]{Canyu Wang} received the M.S. degree in mechanical engineering from the Beijing Institute of Technology, Beijing, China, in 2015.He is currently an Engineer with the China North Vehicle Research Institute, and is pursuing the Ph.D. degree in vehicle engineering with the Beijing Institute of Technology. His current research interests include the drive control of tracked vehicles, and learning theory and methods for dynamics and control.
        \end{IEEEbiography}
        
		\begin{IEEEbiography}[{\includegraphics[width=1in,height=1.25in,clip,keepaspectratio]{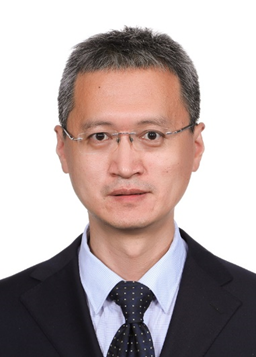}}]{Junqiang Xi} received the B.S. degree in automotive engineering from the Harbin Institute of Technology, Harbin, China, in 1995, and the Ph.D. degree in vehicle engineering from the Beijing Institute of Technology (BIT), Beijing, China, in 2001. In 2001, he joined the State Key Laboratory of Vehicle Transmission, BIT. During 2012–2013, he did research as an Advanced Research Scholar in Vehicle Dynamic and Control Laboratory, Ohio State University, Columbus, OH, USA. He is currently a Professor and Director of the Automotive Research Center at BIT. His research interests include vehicle dynamics and control, powertrain control, mechanics, intelligent transportation systems, and intelligent vehicles.
		\end{IEEEbiography}
        
		
		%\vfill
		

	\end{CJK}
\end{document}